%% file: ms.tex
\documentclass[journal]{IEEEtran}
\pdfoutput=1

\usepackage{graphicx} % for pdf, bitmapped graphics files
\usepackage{epstopdf}

\usepackage{amsmath} % assumes amsmath package installed
\usepackage{amsthm}  % For special theorem style
\usepackage{amsfonts}
\usepackage[dvipsnames]{xcolor}
\usepackage{tikz}
\usepackage{pgfplots}
\usetikzlibrary{shapes,arrows}
\usetikzlibrary{backgrounds}
\usepackage{multirow}
\usepackage[keeplastbox]{flushend}
\usepackage[utf8]{inputenc}    % utf8 support
\usepackage[T1]{fontenc}       % code for pdf file
\usepackage[USenglish]{babel}  % language support

\usepackage{url}
\usepackage[numbers]{natbib}
\usepackage{letltxmacro}
\LetLtxMacro{\autocite}{\citep}
\LetLtxMacro{\textcite}{\citet}

\usepackage{booktabs}
\usepackage{tabularx}
\newcommand{\otoprule}{\midrule[\heavyrulewidth]}

\usepackage{pgfplots}
\pgfplotsset{compat=1.9}  % Prevent warning, pgf running in backwards compatibility mode anyway%\usetikzlibrary{external}                       %% Create pdf figures from TikZ. Use PDFTeXify ...
\usepackage[capitalize]{cleveref}
\Crefname{figure}{Figure}{Figures}
\crefname{equation}{}{}
\Crefname{equation}{Equation}{Equations}
\usepackage{subcaption}
\usepackage{xstring}
\usepackage{xparse}
\usepackage{siunitx}
\usepackage[nolist,nohyperlinks]{acronym}
\makeatletter
\newcommand*{\org@overidelabel}{}
\let\org@overridelabel\@verridelabel
\@ifpackagelater{acronym}{2015/03/21}{% v1.41
  \renewcommand*{\@verridelabel}[1]{%
    \@bsphack
    \protected@write\@auxout{}{\string\AC@undonewlabel{#1@cref}}%
    \org@overridelabel{#1}%
    \@esphack
  }%
}{% older versions
  \renewcommand*{\@verridelabel}[1]{%
    \@bsphack
    \protected@write\@auxout{}{\string\undonewlabel{#1@cref}}%
    \org@overridelabel{#1}%
    \@esphack
  }%
}
\makeatother

\theoremstyle{plain}
\newtheorem{definition}{Definition}
\theoremstyle{remark}\newtheorem{remarkenv}{Remark}        %% remarks
\newenvironment{remark}{\begin{remarkenv}}%
	{\hfill$\lozenge$\end{remarkenv}}            %% end remark with a lozenge

\graphicspath{ {figures/} }

\newlength\figurewidth
\newlength\figureheight
\newlength\venncircle
\newlength\objectwidth

\usetikzlibrary{arrows,positioning}
\usetikzlibrary{arrows.meta}
\definecolor{TNOlightgray}{RGB}{226,226,226}
\definecolor{abstractclass}{RGB}{255, 255, 255}
\definecolor{scenariocategory}{RGB}{226,226,226}
\definecolor{category}{RGB}{226,226,226}
\definecolor{scenario}{RGB}{255, 255, 255}
\definecolor{otherclass}{RGB}{255, 255, 255}
\tikzstyle{tag}=[text height=.8em, text depth=.1em, font=\small\sffamily, rounded corners=0.2em, fill=TNOlightgray, node distance=9em, text width=7em, align=center]
\tikzstyle{tag wide}=[tag, text width=12em]
\tikzstyle{tagarrow}=[->, line width=0.75mm, color=TNOlightgray]
\tikzstyle{object}=[draw, text width=\objectwidth-.5em, align=left, line width=1pt, minimum width=\objectwidth, anchor=north west, node distance=3pt]
\newcounter{tagbcounter}
\newcounter{tagccounter}
\newcommand{\taga}[1]{
	\node[tag wide](taga){#1};
	\node[coordinate, below of=taga, node distance=1.2em](helper){};
}
\newcommand{\tagb}[3]{
	\node[tag, below of=taga, node distance=3em, xshift=#3](tagb#2){#1};
	\draw[tagarrow] (taga) -- (helper) -| (tagb#2);
	\node[coordinate, xshift=1em](tagb#2 helper) at (tagb#2.south west) {};
}
\newcommand{\tagc}[4]{
	\node[tag, below of=tagb#4, node distance=#3em, xshift=2em](tagc#2){#1};
	\draw[tagarrow] (tagb#4 helper) |- (tagc#2);
}
\ExplSyntaxOn
\NewDocumentCommand{\tree}{O{} m m}{%
	\begin{tikzpicture}
	\taga{#2}
	\setcounter{tagbcounter}{0}
	\seq_set_split:Nnn \arg { ; } { #3 }
	\seq_map_inline:Nn \arg {
		\seq_set_split:Nnn \argb {,} {##1}
		\seq_pop_left:NN \argb \argl
		\tagb{\argl}{\arabic{tagbcounter}}{\arabic{tagbcounter}*10em-\seq_count:N \arg *5em+5em}
		\setcounter{tagccounter}{0}
		\seq_map_inline:Nn \argb {
			\stepcounter{tagccounter}
			\tagc{####1}{\arabic{tagccounter}}{2*\arabic{tagccounter}}{\arabic{tagbcounter}}
		}
		\stepcounter{tagbcounter}
	}
	#1
	\end{tikzpicture}
}
\ExplSyntaxOff

\newlength\blockwidth
\newlength\blockheight
\newlength\blockx
\newlength\blocky
\newlength\legendwidth
\tikzstyle{class}=[draw, text width=\blockwidth-.5em, align=center, minimum height=\blockheight, line width=1pt, minimum width=\blockwidth]
\tikzstyle{aggregation}=[-{Diamond[width=8pt, length=12pt, fill=white]}, line width=1pt]
\tikzstyle{falls into}=[->, line width=1pt]
\tikzstyle{superclass}=[-{Triangle[width=8pt, length=12pt, fill=white]}, line width=1pt]

\newcommand{\amplitude}{A}
\newcommand{\attrtstart}{start event}
\newcommand{\attrtend}{end event}
\newcommand{\comprises}{\ni}

\newcommand{\distancecondition}{d_{\mathrm{v,p}}}
\newcommand{\duration}{T}
\newcommand{\east}{x}
\newcommand{\egosub}{ego}
\newcommand{\egoeast}{\east_{\mathrm{\egosub}}}
\newcommand{\egonorth}{\north_{\mathrm{\egosub}}}
\newcommand{\egospeed}{v_{\mathrm{\egosub}}}
\newcommand{\egoheading}{\head_{\mathrm{\egosub}}}
\newcommand{\function}{f}
\newcommand{\hasone}{$1$}
\newcommand{\hastwo}{$2$}
\newcommand{\hasn}{$0,1,\ldots$}
\newcommand{\head}{\phi}
\newcommand{\includes}{\supseteq}
\newcommand{\inputsystem}{u}
\newcommand{\north}{y}
\newcommand{\origin}{O}
\newcommand{\parameter}{\theta}
\newcommand{\parametera}{a}
\newcommand{\parameterb}{b}
\newcommand{\pedsub}{ped}
\newcommand{\pedeast}{\east_{\mathrm{\pedsub}}}
\newcommand{\pednorth}{\north_{\mathrm{\pedsub}}}
\newcommand{\pedheading}{\head_{\mathrm{\pedsub}}}
\newcommand{\scenario}{S}

\newcommand{\scenariocategory}{\mathcal{C}}
\newcommand{\scenariocategorya}{\scenariocategory_{1}}
\newcommand{\scenariocategoryb}{\scenariocategory_{2}}

\newcommand{\slope}{s}
\newcommand{\state}{z}
\newcommand{\statedot}{\dot{\state}}
\newcommand{\stateinit}{\state_{0}}
\renewcommand{\time}{t}
\newcommand{\timeinit}{\time_{0}}

\begin{acronym}[AAAAAAAA]
	\acro{av}[AV]{Automated Vehicle}
	\acroindefinite{av}{an}{an}
	\acro{odd}[ODD]{Operational Design Domain}
	\acroindefinite{odd}{an}{an}
	\acro{oof}[OOF]{Object-Oriented Framework}
	\acroindefinite{oof}{an}{an}
\end{acronym}

\begin{document}
\date{}

\title{Towards an Ontology for Scenario Definition for the Assessment of Automated Vehicles: An Object-Oriented Framework}

\author{Erwin~de~Gelder$^{1,2,*}$,
	    Jan-Pieter~Paardekooper$^{1,3}$,
	    Arash~Khabbaz~Saberi$^{1}$,
	    Hala~Elrofai$^{1}$,
	    Olaf~Op~den~Camp$^{1}$,
	    Steven~Kraines$^{4}$,
	    Jeroen~Ploeg$^{5,6}$,
	    Bart~De~Schutter$^{2}$%
\thanks{$^1$ TNO, Dept.\ of Integrated Vehicle Safety, Helmond, The Netherlands}%
\thanks{$^2$ Delft University of Technology, Delft Center for Systems and Control, Delft, The Netherlands}%
\thanks{$^3$ Radboud University, Donders Institute for Brain, Cognition and Behaviour, Nijmegen, The Netherlands}%
\thanks{$^4$ Symphony Co., Tokyo, Japan}%
\thanks{$^5$ 2getthere, Utrecht, The Netherlands}%
\thanks{$^6$ Eindhoven University of Technology, Dept.\ of Mechanical Engineering Dynamics and Control group, Eindhoven, The Netherlands}%
\thanks{$^*$ Corresponding author: \textit{erwin.degelder@tno.nl}}}%

%%%%%%%%%%%%%%%%%%%%%%%%%%%%%%%%%%%%%%%%%%%%%%%%%%%%%%%%%%%%%%%%%%%%%%%%%%%%%%%%
\maketitle
\input{sections/abstract}
\acresetall

%%%%%%%%%%%%%%%%%%%%%%%%%%%%%%%%%%%%%%%%%%%%%%%%%%%%%%%%%%%%%%%%%%%%%%%%%%%%%%%%
\input{sections/introduction}
\input{sections/background}
\input{sections/definitions}
\input{sections/framework}
\input{sections/example}
\acresetall
\input{sections/conclusion}
\appendices
\crefalias{section}{appendix}
\input{sections/nomenclature}
\input{sections/acknowledgement}

\bibliographystyle{IEEEtranN}
\bibliography{ontologybib}

\end{document}

%% file: sections/abstract.tex
\begin{abstract}
	% Explain issue: assessment is important
	The development of new assessment methods for the performance of automated vehicles is essential to enable the deployment of automated driving technologies, due to the complex operational domain of automated vehicles.
	% Introduce scenarios (for assessment)
	One contributing method is scenario-based assessment in which test cases are derived from real-world road traffic scenarios obtained from driving data.
	% Problem -> definition needed for scenario
	Given the complexity of the reality that is being modeled in these scenarios, it is a challenge %to know when enough information is captured in the scenario for defining an effective test case. Depending on the situation, the critical parameters that are needed in a particular scenario differ. This fact makes it difficult 
	to define a structure for capturing these scenarios.
	An intensional definition that provides a set of characteristics that are deemed to be both necessary and sufficient to qualify as a scenario assures that the scenarios constructed are both complete and intercomparable.
	
	% Solution: concrete definitions
	In this article, we develop a comprehensive and operable definition of the notion of scenario while considering existing definitions in the literature.
	This is achieved by proposing an object-oriented framework in which scenarios and their building blocks are defined as classes of objects having attributes, methods, and relationships with other objects.
	%We also introduce the concept of a scenario category that enables the categorization of scenarios in terms of the categories of their typical components.
	% Benefits of our approach
	The object-oriented approach promotes clarity, modularity, reusability, and encapsulation of the objects. 
	We provide definitions and justifications of each of the terms.
	Furthermore, the framework is used to translate the terms in a coding language that is publicly available.
	% Example
	%An example \cstartf of a scenario in which a vehicle approaches a pedestrian crossing \cendf illustrates that the presented \cstartb framework \cendb is applicable for scenario-based assessment of automated vehicles.
\end{abstract}

%% file: sections/introduction.tex
\section{Introduction}
\label{sec:introduction}

% Introduce scenario-based approach and why scenario definition is important.
An essential aspect in the development of \acp{av} is the assessment of quality and performance aspects of the \acp{av}, such as safety, comfort, and efficiency \autocite{bengler2014threedecades, wachenfeld2016release, Helmer2017safety, stellet2015taxonomy, gietelink2006development, putz2017pegasus, roesener2017comprehensive, riedmaier2020survey}.
For legal and public acceptance of \acp{av}, a clear definition of system performance is important, as well as quantitative measures for system quality. 
According to \autocite{wachenfeld2016release}, traditional methods for evaluating driver assistance systems, such as \autocite{response2006code, ISO26262}, cannot sufficiently assess quality and performance aspects of \iac{av}, because they would require too many resources. 
A scenario-based approach could be a viable way to perform the \ac{av} assessment \autocite{putz2017pegasus, elrofai2018scenario, riedmaier2020survey}. 
% Explain importance/relevance
For a scenario-based assessment, proper specification of scenarios is crucial because 
\begin{itemize}
	\item scenarios provide the basis and justification for the tests used for the scenario-based assessment \autocite{stellet2015taxonomy, aparicio2013pre, ulbrich2015, geyer2014, putz2017pegasus, zofka2015datadrivetrafficscenarios},
	\item it helps to arrive at an unambiguous description of scenarios that is crucial for providing standardized, repeatable, and reproducible tests \autocite{aparicio2013pre},
	\item standardized descriptions of scenarios can be more easily compared and classified automatically \autocite{degelder2019scenariocategories},
	\item properly specified scenarios are the basis for evaluating the coverage of the assessment \autocite{putz2017pegasus}, and
	\item properly specified scenarios enable us to translate the result of a test into an assessment of the \ac{av} performance with regards to a particular \ac{odd} \autocite{weber2019framework, gyllenhammar2020towards}.
\end{itemize}

% What is currently missing
Although the notion of scenario is frequently used in the context of automated driving \autocite{gietelink2006development, ebner2011identifying, hulshof2013autonomous, xiong2015orchestration, zofka2015datadrivetrafficscenarios, putz2017pegasus, roesener2017comprehensive, shao2019evaluating, neurohr2021criticality}, only rarely is an explicit definition actually given. 
Furthermore, even those definitions are unclear because of ambiguities and the use of other undefined terms. 
From the implementation perspective, describing scenarios unambiguously becomes more important given the many simulators that are recently being introduced \autocite{rosique2019systematic, wimmer2019toward, chao2020survey, kaur2021survey, nehemiah2021building}.
To this end, there are several file formats and methods for defining scenarios for the assessment of \acp{av}, such as OpenSCENARIO \autocite{openscenario} and CommonRoad \autocite{althoff2017CommonRoad}.
Because the focus of these implementations is on scenarios that can be simulated, these implementations describe scenarios at a quantitative level and, consequently, they do not provide concepts for a qualitative description of a scenario.
Furthermore, these implementations and other object-oriented approaches used in the field of the assessment of \acp{av} \autocite{tsai2003scenario, utting2012taxonomy, zofka2016testing, wittmann2017method} mostly lack the definitions and justifications of each of the terms.

% Our proposal
In this work, as a starting point for developing a full ontology of scenarios, we propose a novel \acf{oof} that addresses the aforementioned shortcomings.
To avoid ambiguities in the definitions, we provide intensional definitions for concepts corresponding to scenarios and all of their essential building blocks (such as activities, actors, and events). 
These intentional definitions give the meaning of the concepts by specifying necessary and sufficient conditions for when the concepts should be used.
We base the definitions of each of the components on definitions that are commonly used in the field of the safety assessment of \acp{av} \autocite{geyer2014, ulbrich2015, catapult2018musicc, catapult2018regulating, sigsim2019glossary, openscenario}. 
While being broadly consistent with existing definitions \autocite{geyer2014, ulbrich2015, elrofai2016scenario}, this framework aims to be sufficiently explicit to enable the formalization of a scenario description. 
More specifically, because we give the characteristics of the concepts corresponding to scenarios and specify how those concepts interrelate, we can define the scenario components as objects of classes having attributes, methods, and relationships with objects that are members of other classes.
In addition to the definition of a scenario, we introduce the concept of a \emph{scenario category} that is used to qualitatively describe scenarios, i.e., an abstraction of a scenario. 
Scenario categories enable the categorization of scenarios in terms of the categories of their typical components.
The presented \ac{oof} provides explicit guidelines for the construction of scenario descriptions that are able to effectively assess the \ac{av} performance.

% Benefits
The proposed approach brings several benefits.
First, we provide concepts for a qualitative description of a scenario, which is useful because it enables to classify scenarios and to interpret scenarios. 
Second, the \ac{oof} allows for reusing and maintaining (the building blocks of) a scenario as well as performing operations on and interacting with (the building blocks of) a scenario.
Third, our framework is supported with the definitions and justifications of each of the concepts.
Fourth, the framework enables the translation of the concepts and their relationships into object-oriented code.
This, in turn, is used to describe scenarios in a coding language that can be understood by various software agents, such as simulation tools, and that can be ported to already available formats like OpenSCENARIO \cite{openscenario}.

% Results
To illustrate how to use the presented \ac{oof}, we have implemented the framework in a coding language that is publicly available at \url{https://github.com/ErwindeGelder/ScenarioDomainModel}\footnote{As a coding language, Python is used. The code implementation also contains more methods than presented in this article.}.
This link contains real-life applications of the presented \ac{oof}, such as describing scenarios extracted from data \autocite{degelder2020scenariomining}.
The framework is also used as a schema for a database system for storing scenarios and scenario categories.
Such a database can be used to perform scenario-based assessment of \acp{av}\footnote{An illustration of such an assessment is publicly available at \url{https://github.com/ErwindeGelder/ScenarioDomainModel}.} \autocite{degelder2021risk}.
To further illustrate the use of the \ac{oof}, this article provides an example with a real-world case  in which a vehicle approaches a pedestrian crossing.
The proposed \ac{oof} provides a first step towards an ontology \autocite{siricharoen2009ontology} for scenarios for the assessment of \acp{av}. In a subsequent study, the formalized concepts presented in this article will be used to design an ontology with logical constraints that enable a computer to perform reasoning on scenarios.

% Outline
The outline of the article is as follows. In \cref{sec:background}, we explain why \iac{oof} is useful and what the context is. 
We define the notions of \emph{scenario}, \emph{event}, \emph{activity}, and \emph{scenario category} in \cref{sec:definitions}. 
The \ac{oof} that formalizes these definitions is presented in \cref{sec:oo framework}. 
In \cref{sec:example}, an application example is provided to illustrate the use of the framework with a real-world scenario. 
The article is concluded in \cref{sec:conclusion}.

%% file: sections/background.tex
\section{Background}
\label{sec:background}

In \cref{sec:why oo framework}, we explain why we want to present \iac{oof} for describing scenarios and  scenario categories. 
\Cref{sec:context} provides information on the context for which we want to define scenarios.

\subsection{Why an  object-oriented framework?}
\label{sec:why oo framework}

According to \textcite{johnson1988designing}, \iac{oof} is a ``set of classes that embodies an abstract design for solutions to a family of related problems.''
The object orientation is used for ``a representation, modeling, and abstraction formalism'' \autocite{wegner1990concepts}, which is why it is considered ``not only useful but also fundamental'' \autocite{wegner1990concepts}.
In addition, \textcite{patridge2005business} notes that object-oriented modeling can provide a bridge from traditional entity-relation-based data modeling to data modeling that is fully grounded in a formalized ontology.
\Iac{oof} offers the following benefits:
\begin{itemize}
	\item \emph{Clarity:} It provides ``a common vocabulary for designers to communicate, document, and explore design alternatives'' \autocite{gamma1993design}.
	\item \emph{Modularity:} By decomposing a scenario into components, the complexity of a scenario itself is reduced. 
	Thus, ``modularity makes it easier to understand the effect of changes'' \cite{johnson1988designing}.
	\item \emph{Reusability:} \Iac{oof} promotes reusability \autocite{snyder1986encapsulation, meyer1987reusability, johnson1988designing}. 
	For example, if two classes share certain procedures and/or properties, these procedures and/or properties could be provided by a so-called superclass from which these two classes inherit the procedures and properties, such that these procedures and properties need to be defined only once.
	\item \emph{Encapsulation:} Encapsulation assures ``that compatible changes can be made safely, which facilitates program evolution and maintenance'' \autocite{snyder1986encapsulation}.
	\item \emph{Possibility to translate to object-oriented programming languages:} As the framework consists of a set of classes, it can be directly used in an object-oriented coding language. 
	The framework then specifies the relationships between the different classes and provides information on the properties of a class and the possible values.
\end{itemize}

\subsection{Context of a scenario}
\label{sec:context}

Because the notion of scenario is used in many different contexts outside of the domain of road traffic, a wide diversity in definitions of this notion exists (for an overview, see \autocite{vannotten2003updated, bishop2007scentechniques}). 
Therefore, it is reasonable to assume that ``there is no [generally] `correct' scenario definition'' \autocite{vannotten2003updated}.
As a result, to define the notion of scenario, it is important to consider the context in which it will be used. 

In this article, the context of a scenario is the assessment of \acp{av}, where \acp{av} refer to vehicles equipped with a driving automation system\footnote{According to \autocite{sae2018j3016}, a driving automation system is ``the hardware and software that are collectively capable of performing part or all of the dynamic driving task on a sustained basis. 
This term is used generically to describe any system capable of level 1-5 driving automation.'' Here, level 1 driving automation refers to ``driver assistance'' and level 5 refers to ``full driving automation''. For more details, see \autocite{sae2018j3016}.}. 
It is assumed that the assessment methodology uses scenario-based test cases. 
The ultimate goal is to build a database with all relevant scenarios that \iac{av} has to cope with when driving in the real world \autocite{putz2017pegasus}. 
Hence, a scenario should be a description of a potential use case of \iac{av}. 

%% file: sections/definitions.tex
\section{Definitions}
\label{sec:definitions}

One of the main reasons to introduce \iac{oof} is to enable sharing of knowledge between researchers, developers, and users. 
Therefore, it is important that the terms we use are clearly defined. 
When presenting our \ac{oof} in \cref{sec:oo framework}, we will formalize the terms such that they can be used by software agents.
In this section, we define the terms \emph{scenario}, \emph{event}, \emph{activity}, and \emph{scenario category}, thereby providing insight into the terms used in the next section.
We aim to provide intensional definitions that are in accordance with the common use of these terms in the literature and to provide clarity on what are the necessary and sufficient conditions for when the term should be used.

We first define the concept of a scenario in \cref{sec:scenario}. 
Next, we define two important components of a scenario: events and activities, in \cref{sec:event,sec:activity}, respectively. 
Lastly, we present the definition of a scenario category in \cref{sec:scenario category}.
Each of the \cref{sec:scenario,sec:event,sec:activity,sec:scenario category} starts with background information. 
Next, we draw conclusions that lead to our proposed definition of the corresponding term. 
After proposing a definition, each section finishes with remarks and implications of the proposed definition.
For the definitions provided in \cref{sec:scenario,sec:event,sec:activity,sec:scenario category}, use is made of the terms listed in \cref{tab:nomenclature}. 
The definitions in \cref{tab:nomenclature} are mostly based on literature; see \cref{sec:nomenclature} for more details.

\begin{table}[t]
	\caption{Terms and definitions that are used in \cref{sec:definitions}. 
		For more details, see \cref{sec:nomenclature}.}
	\label{tab:nomenclature}
	\begin{tabularx}{\linewidth}{lX}
		\toprule
		Term & Definition \\ \otoprule
		Ego vehicle & Vehicle from which the world is perceived and/or vehicle that must perform a certain task during a test \\
		Physical element & Object that exists in the three-dimensional space \\
		%Static physical thing & Physical thing that does not experience a (relevant) change during a scenario \\
		% & Note: All relevant static physical things together form the static environment. \\
		%Dynamic physical thing & Physical thing that experiences a (relevant) change during a scenario \\
		Actor & Physical element that experiences change \\
		& Note: An actor is a physical element, but a physical element is not necessarily an actor. \\
		Static environment & Part of the environment that does not change \\
		Dynamic environment & Part of the environment that does change and that comprises all actors \\
		Act & Combination of an actor and an activity \\
		State variables & Description of the present configuration of a system that can be used to determine the future response, given the excitation inputs and the equations describing the dynamics \\	
		State vector & Vector containing all $n$ state variables \\
		Model & Differential and algebraic equations that describe the dynamics \\
		Mode & Period in which a system does not exhibit a sudden change in an input, a model parameter, or the model \\	
		\bottomrule
	\end{tabularx}
\end{table}

\subsection{Scenario}
\label{sec:scenario}

% Definition according to Go and Carroll
\textcite{go2004blind} describe a scenario within the field of system design. They define a scenario as ``a description that contains (1) actors, (2) background information on the actors and assumptions about their environment, (3) actors' goals or objectives, and (4) sequences of actions and events. 
Some applications may omit one of the elements or they may simply or implicitly express it. 
Although, in general, the elements of scenarios are the same in any field, the use of scenarios is quite different.'' 

% Definition according to Geyer et al.
\textcite{geyer2014} describe a scenario within the context of automated driving. 
They use the metaphor of a movie or a storybook for describing a scenario and state that ``a scenario includes at least one situation within a scene including the scenery and dynamic elements. 
However, [a] scenario further includes the ongoing activity of one or both actors.'' 
\textcite{geyer2014} define a scene ``by a scenery, dynamic elements, and optional driving instructions.''
In \autocite{geyer2014}, the meaning of activity is not detailed.

% Definition according to Ulbrich et al.
\textcite{ulbrich2015} define a scenario as ``the temporal development between several scenes in a sequence of scenes. 
Every scenario starts with an initial scene. 
Actions \& events as well as goals \& values may be specified to characterize this temporal development in a scenario. 
Other than a scene, a scenario spans a certain amount of time.'' 
The authors of \autocite{ulbrich2015} state that actions and events link the different scenes. 
A further description of actions and events is not given in \autocite{ulbrich2015}.

% Definition according to Elrofai et al.
Another definition of a scenario in the context of automated driving is given by \textcite{elrofai2016scenario}. 
They define a scenario as ``the combination of actions and maneuvers of the host vehicle in the passive [i.e., static] environment, and the ongoing activities and maneuvers of the immediate surrounding active [i.e., dynamic] environment for a certain period of time.'' 

\textcite{catapult2018regulating} define a scenario as ``a description of a short interaction between an AV and other road users and/or road infrastructure''. 

In a concept paper on OpenSCENARIO 2.0 \autocite{OpenSCENARIO2}, a scenario is defined as ``a `description of the temporal development' of road users (actor entities) defined by their actions, where temporal activation (defining when) `is regulated by' conditional `triggers'. A scenario comprises both scenery and dynamic elements.''

% "Requirements"
As a basis for constructing a comprehensive definition for the concept of scenario, we list the major characteristics contained in the above definitions as follows:
\begin{enumerate}
	\item \textit{A scenario corresponds to a time interval.}
	The aforementioned definitions \autocite{go2004blind, geyer2014, ulbrich2015, elrofai2016scenario} state that a scenario corresponds to a time interval. \textcite{vannotten2003updated} call such a scenario a chain scenario (``like movies''), as opposed to a snapshot scenario, i.e., a scenario that describes the state at a given time instant (``like photos''). 

	% Scenarios consists of one or several events
	\item \textit{A scenario consists of two or more events \autocite{vannotten2003updated, go2004blind, geyer2014, ulbrich2015, kahn1962}.}
	It can be helpful to develop scenarios using events \autocite{bishop2007scentechniques}. 
	Thus, a scenario could be defined as a particular sequence of events or, as \textcite[p.~143]{kahn1962} writes, ``a scenario results from an attempt to describe in more or less detail some hypothetical sequence of events''. 
	Furthermore, \textcite{geyer2014} and \textcite{ulbrich2015} use the notion of event for describing a scenario, although they do not provide a definition of the term \emph{event}. 
	Because a scenario contains at least a start event and an end event, the minimum number of events is two.
	In \cref{sec:event}, we will elaborate on the notion of \emph{event}.

	% Semantically described
	\item \textit{Real-world traffic scenarios are quantitative scenarios.}
	Regarding the nature of the data, a scenario can be either qualitative or quantitative \autocite{vannotten2003updated}. 
	For a real-world traffic scenario to be suitable for simulation purposes, it must be described quantitatively.
	A scenario, however, can also be described qualitatively, such that it is readable and understandable for human experts. 
	Providing a qualitative description of a quantitative scenario has become known as a story-and-simulation approach \autocite{alcamo2001scenarios}. 
	Note that a qualitative description of a scenario does not uniquely define a quantitative scenario.
	A qualitative description can be regarded as an abstraction of the quantitative scenario, see also \cref{sec:scenario category}.

	% Some relevance between events
	\item\textit{The time interval of a scenario contains all relevant events.}
	According to \textcite{geyer2014}, ``the end of a scenario is defined by the first irrelevant situation with respect to the scenario''. In a similar manner, we require that the time interval of a scenario should contain all relevant events. Note that `relevant' is subjective and, therefore, an event is considered to be relevant with respect to the perspective of one or more of the participating actors, often called the ``ego vehicle''.

	% Description of environment
	\item\textit{A scenario includes the description of the environment.}
	A scenario should include the description of the static and dynamic environment.
	Although the description of the static environment is not a general prerequisite of a scenario, this is often included when speaking about traffic scenarios \autocite{geyer2014, ulbrich2015, elrofai2016scenario, ebner2011identifying, althoff2017CommonRoad}.
	The static environment consists of all relevant\footnote{The term `relevant' is subjective and depends on the use of the scenario. The composer of a scenario typically judges whether something might be relevant for the scenario.} physical elements that do not undergo relevant changes with respect to the ego vehicle (s) within the time interval between the start and the end of the scenario. 
	The dynamic environment consists of all relevant actors that undergo changes that are relevant to the ego vehicle(s). 
	For example, the road may be part of the static environment, but if the change in the road temperature is relevant to the ego vehicle(s), the road is part of the dynamic environment.
	
	\item\textit{A scenario includes at least one ego vehicle \autocite{geyer2014, elrofai2016scenario}.}
	Because of the two previously mentioned characteristics, a scenario is required to include at least one ego vehicle.
	Note that an ego vehicle is often regarded as the device under test. In this article, however, this is not necessary because the ego vehicle is just the vehicle whose perspective is used to define what is relevant in the scenario.
		
	% Goals (instead of activities)
	\item\textit{A scenario describes the goals or activities of the actors.}
	Either the activities, the goals, or a combination of activities and goals are required to determine how each actor in a scenario responds to specific events.
	Note that this also holds for the ego vehicle since the ego vehicle is an actor.
	When describing a scenario using real-world data, goals do not need to be given; e.g., \textcite{elrofai2016scenario} mention the activities of the actors rather than the goals. 
	When describing a scenario that an AV has to cope with, however, the ego vehicle's goals (i.e., its driving mission \autocite{geyer2014}) could be specified rather than its activities \autocite{ulbrich2015}. 
	Note that if the activities of an actor are described rather than its goals, an observer might not be able to determine whether the actor has successfully responded to the scenario.
\end{enumerate}

% Definition
Hence, we define a scenario as follows:
\begin{definition}[Scenario]\label{def:scenario}
	A scenario is a quantitative description of the relevant characteristics and activities and/or goals of the ego vehicle(s), the static environment, the dynamic environment, and all events that are relevant to the ego vehicle(s) within the time interval between the first and the last relevant event.
\end{definition}

When applying \cref{def:scenario} in \iac{oof}, it is possible 
to give the ``description'' of a component of a scenario simply by providing a reference to that component.
A reference could be, e.g., the full name of a file, a pointer pointing to a specific part of the computer memory, or an identifier that addresses a specific entry in a database.
The advantage of references is that these parts of the scenario can be exchanged across different scenarios, as these scenarios can use the same references. 
As an example, an OpenSCENARIO file allows to provide a reference to an OpenDRIVE file, which describes a road network \autocite{dupuis2010opendrive}. 
As we will see in \cref{sec:oo framework}, in our proposed framework, a scenario may contain references to physical elements, activities, actors, and events.

\subsection{Event}
\label{sec:event}

As mentioned in \cref{def:scenario}, a scenario consists of events. 
The term event is used in many different fields, e.g.:
\begin{itemize}
	\item In computing \autocite{breu1997towards}, an event is an action or occurrence recognized by software. A common source of events are inputs by the software users. An event may trigger a state transition.
	
	\item In probability theory, an event is an outcome or a set of outcomes of an experiment \autocite{pfeiffer2013concepts}. For example, a thrown coin landing on its tail is an event.
	
	\item In the field of hybrid systems theory, ``the continuous and discrete dynamics interact at `event' or `trigger' times when the continuous state [vector] hits certain prescribed sets in the continuous state space'' \autocite{branicky1998hybridcontrol}. Moreover, ``a hybrid system can be in one of several modes, [...], and the system switches from one mode to another due to the occurrence of events'' \autocite{deschutter2000optimal}.
	
	\item In the ISO~15926-2 standard, an ontology for long-term data integration, access, and exchange is specified in which an event is defined as ``a \emph{possible\_individual}\footnote{``An entity that exists in space and time'' \autocite{batres2007upper}.} with zero extent in time, which means that it occurs at an instant in time'' \autocite{batres2007upper}.
	
	\item In event-based control, a control action is computed when an event is triggered, as opposed to the more traditional approach where a control action is periodically computed \autocite{heemels2012eventcontrol}. 
	In event-based control, the event is triggered at the moment at which the system (is about to) reach a certain threshold.
\end{itemize}

Before providing the definition of an event, the following is concluded about an event, based on the aforementioned literature:

% It is a time instant
\begin{enumerate}
	\item\textit{An event corresponds to a time instant.}
	For the definition of event, we consider a hybrid-systems setting with a linear-time model \autocite{alur1994theory}. Therefore, an event happens at some time instant.
	
	% Event should mark transition of a state from one set to another - mention relation with hybrid control
	\item\textit{An event marks a mode transition or the moment a system reaches a threshold.} 
	A mode transition may be induced by either an abrupt change of an input signal, a change of a parameter, a change in the model, or an external cause. 
	It is also possible that the event marks the moment that a system reaches a threshold.
\end{enumerate}

% Give definition
Hence, we define an event as follows:
\begin{definition}[Event] \label{def:event}
	An event corresponds to a moment at which a mode transition occurs or a system reaches a specified threshold, where the former can be induced by both internal and external causes. 
\end{definition}

% Mention that there are basically two definitions. First definition more suitable for triggers for test cases. Second definition more suitable for observations and inputs.
\Cref{def:event} indicates that the moment of an event can be defined in two different ways: (1) by a mode transition or (2) by the system reaching a threshold. 
The first type could be a mode transition caused by a sudden driver input. An event might also be induced by an external cause, such as an environmental change. 
The second type of event, i.e., related to the system reaching a threshold, is especially useful when describing test scenarios.
For example, consider the ego vehicle approaching a pedestrian that is about to cross the road \autocite{seiniger2015test}. 
Here, the event marks the moment that the distance between the vehicle and pedestrian is less than $\distancecondition$ meters. 
At the moment of this event, the pedestrian starts to cross the road such that the vehicle would impact with the pedestrian if it would not change its speed or direction \autocite{seiniger2015test}.
By using a variable threshold $\distancecondition$, the value is flexible and can be set differently to define multiple scenarios.

For the practical implementation of events, a set of conditions may be specified. In that case, the event occurs at the moment that the conditions are met. In \autocite{openscenario}, an extensive list of possible conditions that can be used to define an event is given. E.g., a condition could be that the distance between the vehicle and the pedestrian is below a certain threshold.

\begin{remark}
	\textcite{geyer2014,ulbrich2015} use the term \emph{scene} to define a scenario.
	Like an event, we consider a scene to correspond to a temporal snapshot of the entire scenario. A scene can be obtained by taking a temporal cross-section of the entire scenario as described in \cref{def:scenario}.
\end{remark}

\subsection{Activity}
\label{sec:activity}

To describe the dynamic environment of a scenario, activities are used. A scenario may also describe the activities of the ego vehicle. 
% [Following sentence removed, because too obvious]
% Therefore, next to events, activities can be seen as the building blocks of a scenario.

Both the terms activity \autocite{geyer2014, elrofai2018scenario, childress2015using, catapult2018musicc, sigsim2019glossary} and action \autocite{geyer2014, ulbrich2015, bagschik2017ontology} are used in the context of automated driving. Although, strictly speaking, the terms action and activity have a slightly different meaning, they are often used for the same purpose:
\begin{itemize}
	\item According to \textcite{ulbrich2015}, actions may be specified for characterizing the temporal development in a scenario.
	\item \textcite{elrofai2018scenario} consider an activity as a building block of the dynamic part of the scenario: ``An activity is a time evolution of state variables such as speed and heading to describe for instance a lane change, or a braking-to-standstill.''
	\item In a glossary for scenario catalog development \autocite{catapult2018musicc}, an activity is defined as ``the state [vector] of an object over an interval of time. An activity starts with an event and ends with another event.''
%	\item According to the Cambridge Dictionary, an activity is defined as ``the doing of something, or something that you are doing, have done, or could do'' \autocite{cambridge2019activity}.
%	\item \textcite{caspersen1985physical} define physical activity ``as any bodily movement produced by skeletal muscles that result in energy expenditure''.
%	\item \textcite{bobick1997movement} simply states that an activity is a ``sequence of movements''. 
	\item In the ISO~15926-2 standard, an activity is defined as ``a \textit{possible\_individual} that brings about change by causing the \textit{event} that marks the \textit{beginning}, or the \textit{event} that marks the \textit{ending} of a \textit{possible\_individual}'' \autocite{batres2007upper}.
\end{itemize}

Before providing the definition of an activity, the following is concluded about an activity based on the aforementioned literature:

\begin{enumerate}
	% Time interval
	\item\textit{An activity corresponds to an inter-event time interval.}
	As opposed to an event, an activity spans a certain time interval.
	Furthermore, the start and the end of an activity are marked by an event.
	
	% Describing the evolution of a state	
	\item\textit{An activity quantitatively describes the time evolution of one or more state variables.}
	Because activities are building blocks of a scenario and a scenario corresponds to a quantitative description, the activities themselves need to be quantitative as well. 
	Therefore, an activity describes the time evolution of one or more state variables, i.e., the trajectory of one or more state variables over an inter-event time interval that corresponds to the activity, where the term state variable is defined in \cref{tab:nomenclature}.
	
	% Performed by something (an actor?)
	\item\textit{An activity is performed by an actor.}
	An activity describes the time evolution of one or more state variables and a state variable corresponds to an actor, e.g., the acceleration of a vehicle. 
	% Note that the actor might be an actor or ego vehicle. 
	%However, it might also be the case that a state does not correspond to an actor. For example, a state describing the weather conditions does not correspond to an actor, so if an activity describes changing weather conditions, the activity is not performed by an actor.
\end{enumerate}

Hence, we define an activity as follows:
\begin{definition}[Activity]
	\label{def:activity}
	An activity is a quantitative description of the time evolution of one or more state variables of an actor between two events.
\end{definition}

As an example, an activity could describe the longitudinal acceleration (or, e.g., speed) during an acceleration or deceleration.
Activities describing the lateral position of a vehicle with respect to the center of the corresponding lane might, e.g., be labeled with ``driving straight'' or ``changing lane''.

\subsection{Scenario category}
\label{sec:scenario category}

% Introduce term scenario category (i.e. qualitative description of scenario)
According to \cref{def:scenario}, a scenario in the context of the performance assessment of an AV needs to be quantitative. 
However, in literature, the term scenario is also used to refer to a collection of scenarios, where this collection of scenarios is described qualitatively. 
For example, in \autocite{USDoT2007precrashscenarios}, a typology of pre-crash scenarios is proposed. 
Here, each of the pre-crash scenarios is an abstraction of many quantitative scenarios. 
Similar studies have been performed to describe scenarios that lead to highway accidents \autocite{adaptive2017d73}, car-cyclist accidents \autocite{opdencamp2014cats}, and car-pedestrian accidents \autocite{lenard2014typical}. 
In \autocite{catapult2017taxonomy}, a taxonomy of scenarios is proposed to qualitatively describe challenging scenarios for automated driving.
In \autocite{neurohr2021criticality}, a distinction is made between so-called functional scenarios, abstract scenarios, logical scenarios, and concrete scenarios. 
These four types of scenario descriptions represent different levels of abstraction with functional scenarios referring to non-formal human-readable scenarios, abstract scenarios referring to formalized declarative descriptions, logical scenarios referring to parameterized scenarios with ranges and distributions of the parameters, and concrete scenarios referring to parameterized scenarios with fixed parameters values.

The aforementioned references \autocite{USDoT2007precrashscenarios, adaptive2017d73, opdencamp2014cats, lenard2014typical, catapult2017taxonomy, neurohr2021criticality} show that the term \emph{scenario} is also used to address qualitative descriptions. 
Since we define a scenario as a quantitative description, we need to introduce a different term to address the qualitative description. 
We propose to use the term \emph{scenario category} to refer to the qualitative description of a scenario. 
A qualitative description can be regarded as an abstraction of a quantitative scenario, whereas a quantitative description can be regarded as a concretization of a qualitative description.

We thus define a scenario category as follows:
\begin{definition}[Scenario category] \label{def:scenario category}	
	A scenario category is a qualitative description of the relevant characteristics and activities and/or goals of the ego vehicle(s), the static environment, and the dynamic environment.
\end{definition}

% What is the purpose of this?
% - Human interpretable
% - Group scenarios that are very similar --> Analysis is easier
% - Completeness
Introducing the concept of scenario categories brings the following benefits:
\begin{itemize}
	\item For a human, it is often easier to interpret a qualitative description than a quantitative description.
	\item Scenarios that have something in common can be grouped together, which enables characterization of types of scenarios and facilitates discussion of scenarios.
	\item The completeness of a set of scenarios can be assessed by considering the completeness of scenario categories (see, e.g., \autocite{hauer2019didwe}) and the completeness of scenarios in each category (see, e.g., \autocite{degelder2019completeness}).
\end{itemize}

% Explain scenario category comprise scenarios
We describe the formal relation between a scenario and a scenario category with the verb ``to comprise'', denoted by $\comprises$. If a specific scenario category $\scenariocategory$ is an abstraction of a specific scenario $\scenario$, then we say that $\scenariocategory$ comprises $\scenario$, or simply $\scenariocategory \comprises \scenario$. 
A given scenario category can comprise multiple scenarios and multiple scenario categories can comprise a specific scenario. 
As a consequence, as opposed to the proposed categorization of scenarios in \autocite{opdencamp2014cats, USDoT2007precrashscenarios, lenard2014typical, lara2019harmonized}, scenario categories do not need to be mutually exclusive. 

% Explain scenario category can include scenario categories
The verb ``to include'' is used to describe the relation between two scenario categories. A scenario category $\scenariocategoryb$ is said to include a scenario category $\scenariocategorya$ if $\scenariocategoryb$ comprises all scenarios that are comprised in $\scenariocategorya$. In that case, we can write $\scenariocategoryb \includes \scenariocategorya$. Thus we have
\begin{equation} \label{eq:scenario category include}
	\scenariocategoryb \includes \scenariocategorya \text{ if } \scenariocategoryb \comprises \scenario\,\,\forall\,\, \{S: \scenariocategorya \comprises \scenario\}.
\end{equation}

% Figure is explained in last few paragraphs, but placed earlier as to appear at the right page of the paper.
\begin{figure*}[t]
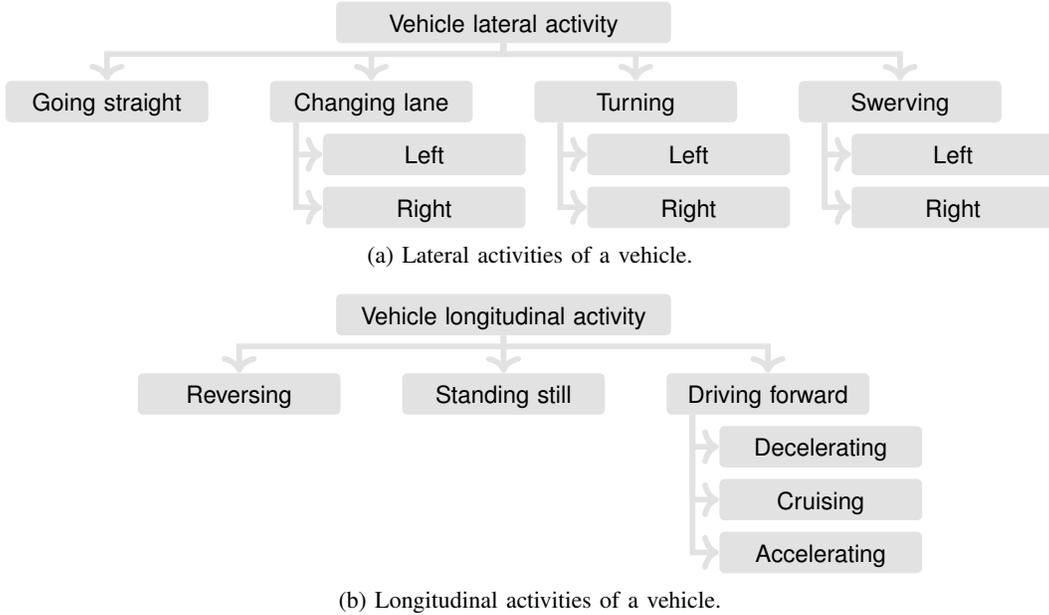

	\centering
	\begin{subfigure}{\linewidth}
		\centering
		\tree{Vehicle lateral activity}{Going straight; Changing lane, Left, Right; Turning, Left, Right; Swerving, Left, Right}
		\caption{Lateral activities of a vehicle.\vspace{1em}}
		\label{fig:tree vehicle lat act}
	\end{subfigure}
	\begin{subfigure}{\linewidth}
		\centering
		\tree{Vehicle longitudinal activity}{Reversing; Standing still; Driving forward, Decelerating, Cruising, Accelerating}
		\caption{Longitudinal activities of a vehicle.}
		\label{fig:tree vehicle long act}
	\end{subfigure}
	\caption{Tags for lateral and longitudinal activities of a vehicle \autocite{degelder2019scenariocategories}. The lateral activity is relative to the lane in which the corresponding vehicle is driving.}
	\label{fig:tree vehicle activities}
\end{figure*}

We propose to provide scenarios and scenario categories with additional information in the form of tags.
A tag is a keyword or a keyphrase that provides extra information on a piece of data \autocite{smith2007tagging}. 
For example, items in a database can contain some tags that enable users to quickly retrieve several items that share a certain characteristic described by a tag \autocite{craft2004tagging}. 
The use of these tags brings several benefits:
\begin{itemize}
	\item The tags of a scenario can be helpful in determining which scenario categories do and do not comprise the scenario.
	\item It is easy to select scenarios from a scenario database or a scenario library by using tags or a combination of tags.
\end{itemize}

There is a balance between having generic scenario categories --- and thus a wide variety among the scenarios comprised by the scenario category --- and having specific scenario categories without much variety among the scenarios comprised by the scenario category. 
For some systems, one may be interested in a very specific set of scenarios, while for another system one might be interested in a set of scenarios with a high variety.
To accommodate this, tags can be structured in hierarchical trees \autocite{molloy2017dynamic}.
The different layers of the trees can be regarded as different abstraction levels \autocite{Bonnin2014}. 

\Cref{fig:tree vehicle activities} shows two examples of trees of tags taken from \autocite{degelder2019scenariocategories}. 
These tags describe possible activities of a vehicle, i.e., the lateral motion control (via steering) and longitudinal motion control (via acceleration and deceleration). 
The tags may refer to the objective of an actor in case no activities are defined. 
For example, a test case in which the ego vehicle's objective is to make a left turn, the tags ``Turning'' and ``Left'' are applicable. 
Note that tags may be used not only to classify vehicle behavior, but also traffic and environment situations, e.g., ``cut-in'' or ``heavy rain''.

%% file: sections/framework.tex
\section{Object-oriented framework for scenarios}
\label{sec:oo framework}

We have already explained the use of \iac{oof} in \cref{sec:why oo framework}. 
In this section, we present our \ac{oof} for scenarios for the assessment of \acp{av}. 
The overview of the framework is formally represented through class diagrams that are briefly presented in \cref{sec:class diagram}. 
Next, \cref{sec:domain scenario category} explains how a scenario category is formally represented in our framework. 
Similarly, in \cref{sec:domain scenario}, we describe how a scenario is formally represented. 
The \ac{oof} can be implemented straightforwardly in object-oriented languages such as C++ and Python, since these languages support the definition of classes, the instantiation of objects from those classes, and concepts such as inheritance and aggregation.
An actual implementation of the \ac{oof} in a coding language is publicly available at \url{https://github.com/ErwindeGelder/ScenarioDomainModel}. 
This link also contains tutorials for the technical application of the \ac{oof}.

\subsection{Class diagrams}
\label{sec:class diagram}

\begin{figure*}
	\centering
	\input{figures/class_overview.tikz}
	\caption{Class-level relationships of most classes of our \acf{oof}.}
	\label{fig:class overview}
\end{figure*}
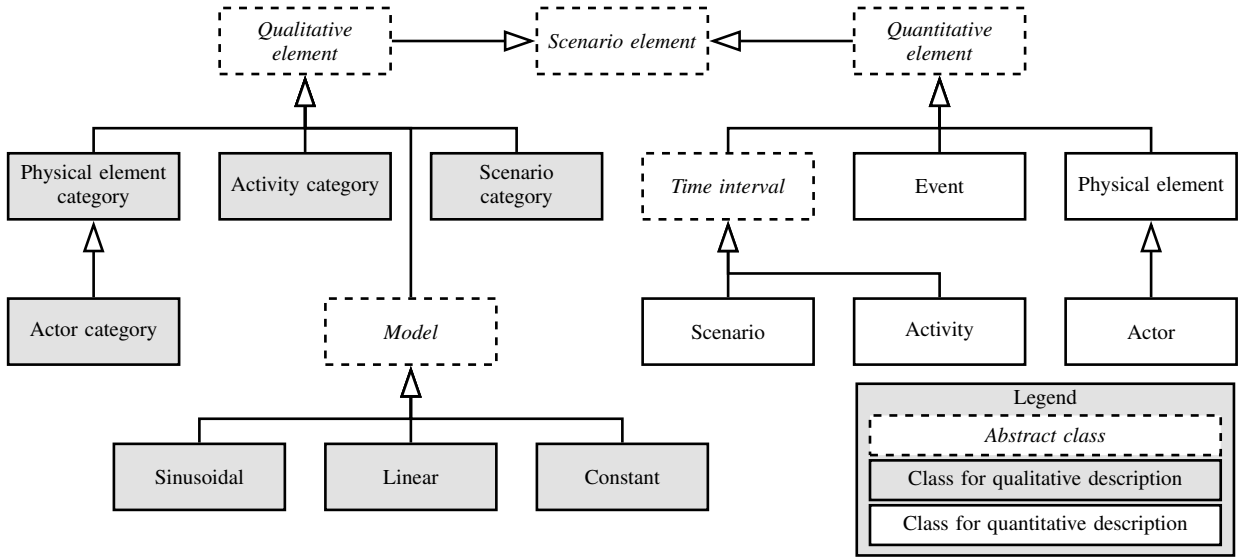

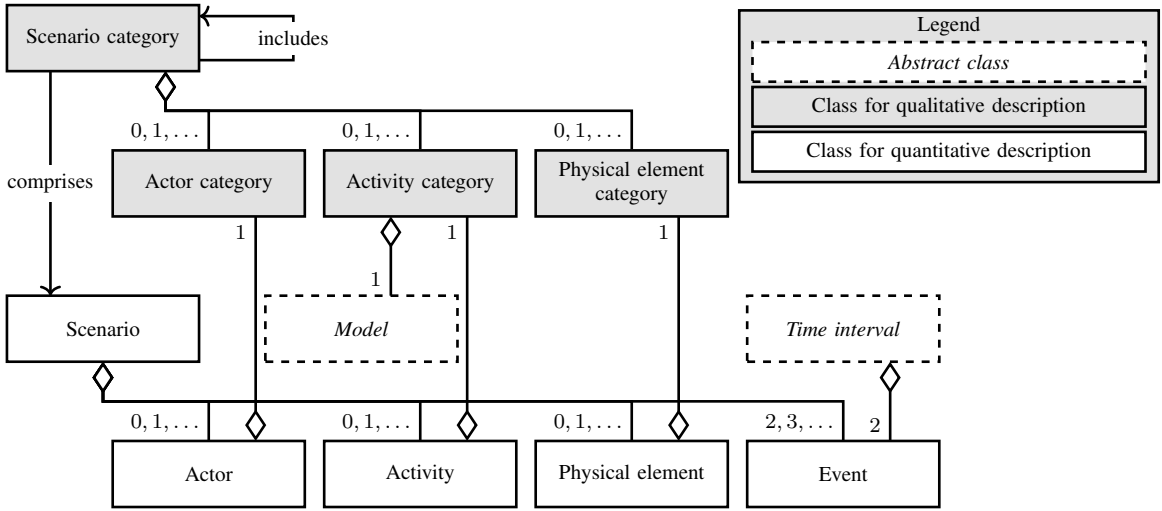
\begin{figure*}[t]
	\centering
	\input{figures/class_relations.tikz}
	\caption{Instance-level relationships of most classes of our \acf{oof}.}
	\label{fig:class relations}
\end{figure*}

In \cref{fig:class overview,fig:class relations}, the gray blocks represent the classes\footnote{In the remainder of this paper, when referring to (an instance of) a class, italic font is used. Additionally, class names start with capital letters and instance names with lowercase letters.} that are used to describe a scenario category according to \cref{def:scenario category} and the white solid blocks represent the classes that are used to describe a scenario according to \cref{def:scenario}. 
The white dashed blocks represent so-called abstract classes. 
Abstract classes cannot be instantiated.
Each class serves as a template for creating objects whereas an object of a particular class is referred to as the instance of that particular class.

\Cref{fig:class overview} shows the class-level relationships while \cref{fig:class relations} shows the instance-level relationships.
In \cref{fig:class overview}, the arrow from, e.g., \textit{Scenario} to \textit{Time interval}, denotes that \textit{Scenario} is a subclass of \textit{Time interval}. 
Therefore, all properties of the \textit{Time interval} are inherited by \textit{Scenario}. 
The arrow with the diamond in \cref{fig:class relations} denotes an aggregation.
This means that, e.g., an \textit{actor}, which is an instance of the \textit{Actor} class, has an \textit{actor category} as an attribute.
Here, the ``\hasone'' at the start of the arrow from \textit{Actor category} to \textit{Actor} indicates that an \textit{actor} has exactly one \textit{actor category}.
Similarly, ``\hastwo'' at the aggregation arrow from \textit{Event} to \textit{Time interval} indicates that a \textit{time interval} contains two \textit{events}, i.e., the events that define the start and the end of the time interval. 
A ``\hasn'' at the start of an aggregation arrow indicates that an object has zero, one, or multiple objects of the corresponding class.
The arrow with the text ``comprises'' and ``includes'' represent methods that are explained in \cref{sec:scenario category}. Here, ``comprises'' can be denoted by $\comprises$ and ``includes'' can be denoted by $\includes$, see \cref{eq:scenario category include}. 

\subsection{Scenario category and its attributes}
\label{sec:domain scenario category}

Because all other classes in \cref{fig:class overview} are subclasses of \textit{Scenario element}, these classes inherit the attributes and procedures of \textit{Scenario element}. 
In our framework, a \textit{scenario element} has a human-interpretable name, a unique ID, and possibly predefined tags that are also interpretable by a software agent. 
So, all other classes in \cref{fig:class overview} also have these attributes.
In addition to these attributes, the \textit{Qualitative element} class has a human-interpretable description.

The static environment is qualitatively described by one or more \textit{physical element categories}.
Because \textit{physical element categories} qualitatively describe the static environment, they contain a human-interpretable description of the physical things they describe.

The ego vehicle(s) and the dynamic environment are qualitatively described by \textit{activity categories} and \textit{actor categories}. 
In line with \cref{def:activity}, \textit{Activity category} includes the state variable(s).
The \textit{Model} that is used to describe the time evolution of the state variable(s) is specified. 
Note that \textit{Model} is an abstract class that serves as a template for different models, such as the three examples shown in \cref{fig:class overview}: \textit{Sinusoidal}, \textit{Linear}, and \textit{Constant}. 
Let $\state(\time)$ denote the state variable at time $\time$, then the \textit{Sinusoidal} model is defined as follows:
\begin{align}
	\statedot(\time) &= \frac{\pi \amplitude}{2\duration} \sin \left( \frac{\pi \left( \time - \timeinit\right)}{\duration} \right),\ \time \in [\timeinit, \timeinit+\duration], \label{eq:sinusoidala} \\
	\state(\timeinit) &= \stateinit. \label{eq:sinusoidalb}
\end{align}
Here, the amplitude ($\amplitude$), duration ($\duration$), initial time ($\timeinit$), and initial state ($\stateinit$) are parameters. 
The \textit{Linear} and \textit{Constant} models are described by the following equations, respectively:
\begin{align}
	\statedot(\time) &= \slope,\ \state(\timeinit) = \stateinit, \label{eq:linear} \\
	\state(\time) &= \stateinit. \label{eq:constant}
\end{align}
The \textit{Linear} model contains three parameters, i.e., the slope ($\slope$), initial time ($\timeinit$), and initial state ($\stateinit$). 
The \textit{Constant} model only has the parameter $\stateinit$.
Since an \textit{activity category} is a qualitative description, the values of the parameters of its \textit{model} are not part of the \textit{activity category}.
Note that this article only considers the models \textit{Sinusoidal}, \textit{Linear}, and \textit{Constant}, but more complex models may be necessary to describe complex behavior. 
More complex models are out-of-scope of this article, but it is straightforward to extend the \ac{oof} with such models.

The \textit{Actor category} is a subclass of \textit{Physical element category} so \textit{Actor category} inherits the properties of \textit{Physical element category}.
In addition, \textit{Actor category} has an attribute that specifies the type of object.
To indicate that an actor is an ego vehicle, the tag ``Ego vehicle'' is added to the list of tags of the \textit{actor category}.

% Scenario category
The \textit{Scenario category} has \textit{physical element categories}, \textit{activity categories}, and \textit{actor categories} as attributes. 
%As with the other classes, a \textit{scenario category} contains a name and may contain predefined tags that describe parts of the scenario that are not described by the other classes.
Another attribute of the \textit{Scenario category} is the list of acts. %\footnote{ In line with the definition of \emph{act} in \cref{sec:act}, for a \textit{scenario category}, an \emph{act} is a combination of \textit{activity categories} and \textit{actor categories}.}. 
These acts describe which actors perform which activities. Note that it is possible that one actor performs multiple activities and that one activity is performed by multiple actors.

% Explain why we have these different classes
The reader might wonder why we introduce the different classes for describing a scenario category, i.e., the gray blocks, instead of only one class for modeling a scenario category. 
The main advantage of the different classes is the reusability of the instances of the classes, because these instances can be exchanged among different \textit{scenario categories}. For example, if two \textit{scenario categories} have the same \textit{actor categories}, we only need to define the \textit{actor categories} once, whereas if the \textit{actor categories} would not be instances of a class but only properties of the scenario category, we would need to define the \textit{actor categories} twice.

\subsection{Scenario and its attributes}
\label{sec:domain scenario}

To distinguish objects that are directly used to compose a \textit{scenario}, these objects are instantiated from subclasses of the \textit{Quantitative element} class.
The class \textit{Scenario} is a subclass of \textit{Time interval} and, therefore, it has \textit{events} that define the start and the end of the scenario.
The \textit{Scenario} also has \textit{physical element}, \textit{activities}, \textit{actors}, and \textit{events} as attributes. 
%The ego vehicle and the dynamic environment are quantitatively described by activities and actors. 
The \textit{physical elements}, \textit{activities}, and \textit{actors} are the quantitative counterparts of the \textit{physical element categories}, \textit{activity categories}, and \textit{actor categories}, just as a \textit{scenario} is the quantitative counterpart of a \textit{scenario category}. 
As with the \textit{Scenario category}, the \textit{Scenario} contains a list of acts that describe which actors perform which activities.

% Static environment
A \textit{physical element} has a \textit{physical element category} and it may have multiple properties that quantitatively define the object, such as its size, weight, color, radar cross section, etc.
Physical elements can be used to define, e.g., the road layout, static weather and lighting conditions, and infrastructural elements.

% Activity
According to \cref{def:activity}, an activity quantitatively describes the evolution of one or more state variables in a time interval. The state variable(s) are defined by the \textit{activity category} that the \textit{activity} has as an attribute. Together with the \textit{Model} that is contained by the \textit{activity category}, the time evolution of the state variable is described by a set of parameters. The values of the parameters are part of the \textit{activity}. 

% Event
Following \cref{def:event}, an \textit{event} contains conditions that describe the threshold or mode transition at the time of the \textit{event}.

% Actor
Similar to a \textit{physical element} and an \textit{activity}, an \textit{actor} has its qualitative counterpart --- an \textit{actor category} --- as an attribute. 
Additionally, the \textit{Actor} contains an initial state vector and a desired state vector that can be used to specify the intent, as attributes.
Describing the intent is especially useful for defining a test scenario in terms of the objective of the ego vehicle rather than its activities.

% Another advantage of the blue blocks -> only one activity category for multiple activities
An advantage of having the qualitative counterparts of the \textit{Physical element}, \textit{Activity}, and \textit{Actor} is that the qualitative description can be reused and exchanged. 
For example, there can be many different braking activities, but there needs to be only one \textit{activity category} for qualitatively defining the braking activity. 
Here, it is assumed that all braking activities are modeled with the same model and that similar tags apply. 
If this is not the case, multiple \textit{activity categories} need to be defined, but the number of \textit{activity categories} will still be substantially lower than the number of \textit{activities}.

%% file: figures/class_overview.tikz
\begin{tikzpicture}
\tikzstyle{every node}=[font=\footnotesize]

% Abstract classes
\node[class, fill=abstractclass, dashed](thing) at (0, 0) {\textit{Scenario element}};
\node[class, fill=abstractclass, dashed](qualitative) at (-1.5\blockx, 0) {\textit{Qualitative element}};
\node[class, fill=abstractclass, dashed](quantitative) at (1.5\blockx, 0) {\textit{Quantitative element}};
\node[class, fill=abstractclass, dashed](timeinterval) at (.5\blockx, \blocky) {\textit{Time interval}};
\node[class, fill=abstractclass, dashed](model) at (-\blockx, 2\blocky) {\textit{Model}};

% Qualitative classes
\node[class, fill=scenariocategory](scencat) at (-0.5\blockx, \blocky) {Scenario category};
\node[class, fill=category](physicalcat) at (-2.5\blockx, \blocky) {Physical element category};
\node[class, fill=category](activitycat) at (-1.5\blockx, \blocky) {Activity category};
\node[class, fill=category](actorcat) at (-2.5\blockx, 2\blocky) {Actor category};
\node[class, fill=category](sinusoidal) at (-2\blockx, 3\blocky) {Sinusoidal};
\node[class, fill=category](linear) at (-\blockx, 3\blocky) {Linear};
\node[class, fill=category](constant) at (0, 3\blocky) {Constant};

% Quantitative classes
\node[class, fill=scenario](scenario) at (.5\blockx, 2\blocky) {Scenario};
\node[class, fill=otherclass](physical) at (2.5\blockx, \blocky) {Physical element};
\node[class, fill=otherclass](activity) at (1.5\blockx, 2\blocky) {Activity};
\node[class, fill=otherclass](event) at (1.5\blockx, \blocky) {Event};
\node[class, fill=otherclass](actor) at (2.5\blockx, 2\blocky) {Actor};

% Superclass arrows
\foreach \fromclass/\toclass in {scencat/qualitative,
								 activitycat/qualitative,
								 physicalcat/qualitative,
								 timeinterval/quantitative,
								 sinusoidal/model,
								 linear/model,
								 constant/model,
								 scenario/timeinterval,
								 activity/timeinterval,
								 event/quantitative,
								 physical/quantitative} {
	\node[coordinate, above of=\fromclass, node distance=-.4\blocky](helper \fromclass){};
	\draw[superclass] (\fromclass) -- (helper \fromclass) -| (\toclass);
}
\foreach \fromclass/\toclass in {qualitative/thing, 
								 quantitative/thing,
								 actorcat/physicalcat,
								 actor/physical} {
	\draw[superclass] (\fromclass) -- (\toclass);
}
\draw[superclass] (model) |- (helper activitycat) -- (qualitative);

% Legend
\node[class, fill=TNOlightgray, minimum height=6.5em, text width=2\blockwidth+1em](legend) at (2\blockx, 2.95\blocky) {};
\node[yshift=2.65em] at (legend) {Legend};
\node[class, fill=abstractclass, minimum height=1.5em, text width=2\blockwidth, yshift=1.3em, dashed](abstract) at (legend) {\textit{Abstract class}};
\node[class, fill=category, minimum height=1.5em, text width=2\blockwidth, yshift=-.4em](category) at (legend) {Class for qualitative description};
\node[class, fill=otherclass, minimum height=1.5em, text width=2\blockwidth, yshift=-2.1em](category) at (legend) {Class for quantitative description};

\end{tikzpicture}

%% file: figures/class_relations.tikz
\setlength{\blockwidth}{7.1em}
\begin{tikzpicture}
\tikzstyle{every node}=[font=\footnotesize]

% Qualitative classes
\node[class, fill=scenariocategory](scenario category) at (.5\blockx,0) {Scenario category};
\node[class, fill=category](actorcategory) at (\blockx, \blocky) {Actor category};
\node[class, fill=category](activitycategory) at (2\blockx, \blocky) {Activity category};
\node[class, fill=abstractclass, dashed](model) at (1.5\blockx+0.25\blockwidth, 2\blocky) {\textit{Model}};
\node[class, fill=category](elementcategory) at (3\blockx, \blocky) {Physical element category};

% Quantitative classes
\node[class, fill=scenario](scenario) at (.5\blockx, 2\blocky) {Scenario};
\node[class, fill=otherclass](actor) at (\blockx, 3\blocky) {Actor};
\node[class, fill=otherclass](activity) at (2\blockx, 3\blocky) {Activity};
\node[class, fill=otherclass](element) at (3\blockx, 3\blocky) {Physical element};
\node[class, fill=otherclass](event) at (4\blockx, 3\blocky) {Event};
\node[class, fill=abstractclass, dashed](timeinterval) at (4\blockx, 2\blocky) {\textit{Time interval}};

% Aggregation arrows for the scenario category
\node[coordinate, below of=scenario category, node distance=-\blocky/2, xshift=\blockwidth/3](helper scenario category){};
\node[coordinate, below of=scenario category, node distance=\blockheight/2, xshift=\blockwidth/3](aggregation scenario category){};
\foreach \class in {actor, activity, element}
{
	\node[coordinate, above of=\class category, node distance=\blockheight/2](helper \class){};  % Needed for later
	\draw[aggregation] (\class category) |- (helper scenario category) -- (aggregation scenario category);
	\node[anchor=south east] at (helper \class) {\hasn};
}

% Aggregation arrow for the model
\foreach \fromclass/\toclass in {model/activitycategory} {
	\node[coordinate, above of=\fromclass, node distance=\blockheight/2, xshift=-\blockwidth/8+\blockx/4](aggregation \fromclass){};
	\node[coordinate, below of=\toclass, node distance=\blockheight/2, xshift=\blockwidth/8-\blockx/4](aggregation \toclass){};
	\draw[aggregation] (aggregation \fromclass) -- (aggregation \toclass);
}
\node[anchor=south east] at (aggregation model) {\hasone};

% Aggregation arrow for scenario
\node[coordinate, below of=scenario, node distance=-\blocky/2](helper scenario){};
\node[coordinate, below of=scenario, node distance=\blockheight/2](aggregation scenario){};
\foreach \class in {actor, activity, element}
{
	\node[coordinate, above of=\class, node distance=\blockheight/2](helper \class){};
	\draw[aggregation] (helper \class) |- (helper scenario) -- (aggregation scenario);
	\node[anchor=south east] at (helper \class) {\hasn};
}
\node[coordinate, above of=event, node distance=\blockheight/2](helper event){};
\draw[aggregation] (helper event) |- (helper scenario) -- (aggregation scenario);
\node[anchor=south east] at (helper event) {$2,3,\ldots$};

% Aggregations for static thing, activity, and actor
\foreach \class in {element, activity, actor}
{
	\node[coordinate, below of=\class category, node distance=\blockheight/2, xshift=\blockwidth/4](category helper){};
	\node[coordinate, above of=\class, node distance=\blockheight/2, xshift=\blockwidth/4](helper){};
	\draw[aggregation] (category helper) -- (helper);
	\node[anchor=north east] at (category helper) {\hasone};
}

% Aggregation for event -> time interval
\node[coordinate, above of=event, node distance=\blockheight/2, xshift=\blockwidth/4](helper event){};
\node[coordinate, below of=timeinterval, node distance=\blockheight/2, xshift=\blockwidth/4](helper timeinterval){};
\draw[aggregation] (helper event) -- (helper timeinterval);
\node[anchor=south east] at (helper event) {\hastwo};

% falls into arrows
\node[coordinate, right of=scenario category, node distance=\blockwidth/2+1pt, yshift=-\blockheight/3](helper1){};
\node[coordinate, right of=scenario category, node distance=\blockwidth/2+1pt, yshift=\blockheight/3](helper2){};
\node[coordinate, right of=helper1, node distance=\blockwidth/2](helper3){};
\node[coordinate, right of=helper2, node distance=\blockwidth/2](helper4){};
\draw[falls into] (helper1) -- (helper3) -- node[fill=white]{includes} (helper4) -- (helper2);
\node[coordinate, above of=scenario, node distance=\blockheight/2, xshift=-.28\blockwidth](helper1){};
\node[coordinate, below of=scenario category, node distance=\blockheight/2, xshift=-.28\blockwidth](helper2){};
\draw[falls into] (helper2) -- node[fill=white, align=center, text width=3.55em]{comprises} (helper1);

% Legend
\node[class, fill=TNOlightgray, minimum height=6.5em, text width=2\blockwidth+1em](legend) at (4.5\blockx, .4\blocky) {};
\node[yshift=2.65em] at (legend) {Legend};
\node[class, fill=abstractclass, minimum height=1.5em, text width=2\blockwidth, yshift=1.3em, dashed](abstract) at (legend) {\textit{Abstract class}};
\node[class, fill=category, minimum height=1.5em, text width=2\blockwidth, yshift=-.4em](category) at (legend) {Class for qualitative description};
\node[class, fill=otherclass, minimum height=1.5em, text width=2\blockwidth, yshift=-2.1em](category) at (legend) {Class for quantitative description};

\end{tikzpicture}

%% file: sections/example.tex
\section{Example: pedestrian crossing}
\label{sec:example}

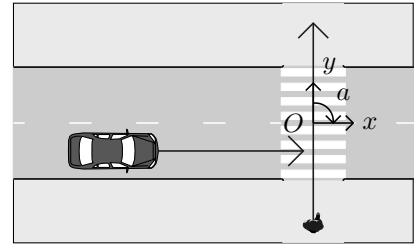
\begin{figure}[t]
	\centering
	\input{figures/example_overview.tikz}
	\caption{Schematic overview of a scenario where both the ego vehicle and a pedestrian are approaching a non-signalized pedestrian crossing. 
		The pedestrian has priority.}
	\label{fig:scenario overview}
\end{figure}

To illustrate the use of the \ac{oof}, we describe a scenario using objects of the classes presented in \cref{sec:oo framework}.
The scenario is schematically shown in \cref{fig:scenario overview}.
The ego vehicle is driving on the right lane of a two-lane road and a pedestrian is walking on a footway that intersects the road the ego vehicle is driving on.
Both the ego vehicle and the pedestrian are initially approaching the pedestrian crossing.
The ego vehicle brakes and comes to a full stop in front of the pedestrian crossing.
While the ego vehicle is stationary, the pedestrian crosses the road using the pedestrian crossing.
When the pedestrian has passed the ego vehicle, the ego vehicle accelerates.
The code of this example is publicly available\footnote{See \url{https://github.com/ErwindeGelder/ScenarioDomainModel}. The repository also contains other examples.}.

\setlength{\figurewidth}{0.6\linewidth}

This particular scenario can be used to formulate a test scenario for the assessment of \iac{av}.
For example, when assessing a pedestrian automatic emergency braking system \autocite{seiniger2015test}, we are interested in the behavior of the system in case the driver or automation system of the ego vehicle does not brake.

We first describe the scenario qualitatively using our proposed framework. 
Next, the scenario is described quantitatively in \cref{sec:example quantitative}. 
In \cref{sec:example test scenario}, we show which objects are reused and which objects are different if we consider an actual test scenario with a crossing pedestrian.

\subsection{Qualitative description of the pedestrian crossing}
\label{sec:example qualitative}

To describe the scenario according to the presented domain model, objects are instantiated from the classes presented in \cref{fig:class overview,fig:class relations}. 
\Cref{fig:example qualitative} shows the objects for describing the scenario qualitatively. 
There are two \textit{actor categories}: one for the ego vehicle and one for the pedestrian. 
Four different \textit{activity categories} are defined: \textit{braking}, \textit{stationary}, \textit{accelerating}, and \emph{walking straight}. 
The \textit{braking}, \textit{stationary}, and \textit{accelerating activity categories} contain the state variable $\egospeed$, i.e., the speed of the ego vehicle, and use the \textit{Sinusoidal} model of \cref{eq:sinusoidala,eq:sinusoidalb}, the \textit{Constant} model of \cref{eq:constant}, and the \textit{Linear} model of \cref{eq:linear}, respectively. 
The \textit{activity category walking straight} has the position of the pedestrian ($\pednorth$) as its state variable and uses the \textit{Linear} model of \cref{eq:linear}.

\begin{figure*}[t]
	\centering
	\input{figures/example_qualitative.tikz}
	\caption{The objects that are used to qualitatively describe the scenario that is schematically shown in \cref{fig:scenario overview}. 
		The first line of each block shows the name (before the double colon) and the class from which the object is instantiated. 
		The following lines show the attributes of the object with the name and value of the attribute before and after the colon, respectively. 
		For the sake of brevity, the unique ID of each object is omitted.}
	\label{fig:example qualitative}
\end{figure*}

The two \textit{actor categories}, the four \textit{activity categories}, and the \textit{physical element category} that represents the crosswalk, are used by the \textit{scenario category}. 
The \textit{scenario category} has four acts. 
The first three acts assign the first three \textit{activity categories} to the ego vehicle.
The last act assigns the \textit{activity category} \emph{walking straight} to the pedestrian.

\subsection{Quantitative description of the pedestrian crossing}
\label{sec:example quantitative}

The objects to describe the scenario quantitatively are shown in \cref{fig:example quantitative}. 
The two \textit{actors} refer to the quantitative counterparts of the \textit{actor categories} in \cref{fig:example qualitative}. 
Initial state vectors are listed for each \textit{actor} using the coordinate frame that is shown in \cref{fig:scenario overview}. 
Since we are describing a real-world scenario, there is no need to define goals or intents for the actors.

\begin{figure*}[t]
	\centering
	\input{figures/example_quantitative.tikz}
	\caption{The objects that are used to quantitatively describe the scenario that is schematically shown in \cref{fig:scenario overview}. 
		For the sake of brevity, the tags and the unique ID of each object are omitted.}
	\label{fig:example quantitative}
\end{figure*}

There are four \textit{events} defined. 
These \textit{events} mark the time instants that define the start and the end of the \textit{activities}.
For simplicity, it is assumed that the start of the scenario occurs at \SI{0}{\second}.

There are four \textit{activities} defined and each of these \textit{activities} refers to its qualitative counterpart.
The \textit{activities} contain the values of the parameters as well as events that mark the start and the end of the \textit{activities}. 
As described by the first \textit{activity} (\emph{ego braking}), the ego vehicle starts with a speed of \SI{8}{\meter\per\second} and brakes in \SI{4}{\second} to come to a full stop. By integrating the sinusoidal function of \cref{eq:sinusoidala} twice, it can be shown that the ego vehicle stops at \SI{4}{\meter} from the center of the pedestrian crossing. 
After waiting for \SI{3}{\second} as described by the second \textit{activity} (\emph{ego stationary}), the ego vehicle accelerates with \SI{1.5}{\meter\per\second\squared} towards a speed of \SI{7.5}{\meter\per\second} as described by the third \textit{activity} (\emph{ego accelerating}). 
The fourth \textit{activity} describes the position and speed of the pedestrian.

The \textit{pedestrian crossing} describes the entire static environment, including the main road the ego vehicle is driving on and the footway the pedestrian is walking on. The example in \cref{fig:example quantitative} shows some properties of the road layout to illustrate how the static environment can be described. Note that, in practice, the quantitative description of the static environment may contain many more facets than the ones mentioned in \cref{fig:example quantitative}. As mentioned in \cref{sec:scenario}, it is possible to refer to another source that contains a description of (part of) the static environment, see, e.g., \autocite{dupuis2010opendrive}. 

The \textit{scenario} has the previously defined \textit{physical element}, \textit{actors}, and \textit{activities} as attributes.
The acts are used to assign the first three \textit{activities} to the ego vehicle and the last \textit{activity} to the pedestrian. 
The \textit{scenario} also has \textit{events} marking the start and the end of the \textit{scenario}.
A different \textit{scenario} can be defined by, e.g., changing the parameter values.
This illustrates that the \textit{scenario category} in \cref{fig:example qualitative} comprises multiple \textit{scenarios}, including the \textit{scenarios} that only differ from the \textit{scenario} in \cref{fig:example quantitative} because of different parameter values.

\subsection{Test scenario of the pedestrian crossing}
\label{sec:example test scenario}

\begin{figure*}[t]
	\centering
	\input{figures/example_test_case.tikz}
	\caption{The objects that, together with the objects \emph{Ego qualitative}, \emph{Pedestrian qualitative}, \emph{Walking straight}, and \emph{Pedestrian crossing qualitative} from \cref{fig:example qualitative} and \emph{Start scenario}, \emph{Pedestrian}, and \emph{Pedestrian crossing} from \cref{fig:example quantitative}, describe a test scenario that is schematically shown in \cref{fig:scenario overview}. 
		For the sake of brevity, the tags and the unique ID of each object are omitted.}
	\label{fig:example test scenario}
\end{figure*}

In this example, we consider a test scenario based on the previously illustrated real-world scenario, see \cref{fig:scenario overview}. 
To describe the test scenario, we reuse the two \textit{actor categories} from \cref{fig:example qualitative} (\emph{ego qualitative} and \emph{pedestrian qualitative}) and the \textit{actor} describing the pedestrian from \cref{fig:example quantitative} (\emph{pedestrian crossing}).
\Cref{fig:example test scenario} shows the other objects that are used to describe this test scenario. 

The \textit{scenario category} only differs from the \textit{scenario category} shown in \cref{fig:example qualitative} in that it does not contain \textit{activity categories} that describe the activity of the ego vehicle.

Two attributes of the quantitative description of the ego vehicle are different. 
First, the initial state vector also includes the speed, denoted by $\egospeed$, at the start of the scenario and the initial position is further away from the pedestrian crossing, such that the ego vehicle's driver or automation system has more time to perceive the pedestrian. 
Second, because there are no activities defined for the ego vehicle, the desired state vector is defined. 
The goal is to reach the point \SI{80}{\meter} in front of the ego vehicle while driving with a speed of $\egospeed=\SI{8}{\meter\per\second}$.

The \textit{event} that marks the start of the walking activity of the pedestrian is triggered if the ego vehicle is \SI{2.5}{\second} away from the center of the footway, assuming that the speed of the ego vehicle is constant. 
In case the ego vehicle drives with a speed of $\egospeed=\SI{8}{\meter\per\second}$, this is at a distance of \SI{20}{\meter}, similar to the scenario described in \cref{fig:example quantitative}.

As with the \textit{scenario category}, the \textit{scenario} does not contain \textit{activities} of the ego vehicle. 
Furthermore, the end event of the scenario is defined differently: now the scenario ends if the ego vehicle either reaches its destination ($\egoeast \geq \SI{20}{\meter}$), collides with the pedestrian, deviates too much from its path ($\egonorth \leq \SI{-2}{\meter}$ or $\egonorth \geq \SI{1}{\meter}$), or takes too long to reach the destination ($\time > \SI{100}{\second}$).

Note that this example considers a pedestrian that crosses the road at a fixed speed (\SI{1}{\meter\per\second}) regardless of the proximity of the ego vehicle.
To model, e.g., the case where the pedestrian notices the ego vehicle and accelerates if a collision is about to happen, an activity can be added that describes the increased speed (e.g., \SI{2}{\meter\per\second}) of the pedestrian.
The start of this activity is at a predefined event with, e.g., the condition $|\egoeast/\egospeed| \leq \SI{1}{\second}$ AND $\pednorth<\SI{0}{\meter}$.

\subsection{Remarks on the example}
\label{sec:remarks example}

% What makes it practical to use in real life?
The example illustrates the benefits of the object-oriented approach for defining a scenario, which are:
\begin{itemize}
	\item clarity regarding the content of the scenario,
	\item modularity, which makes it easy to understand the changes from the real-world scenario in \cref{fig:example quantitative} to the test scenario in \cref{fig:example test scenario}, and
	\item reusability, as is illustrated by the objects that are used more than once.
\end{itemize}
Furthermore, each object listed in \cref{fig:example qualitative,fig:example quantitative,fig:example test scenario} is directly translatable to an object in an object-oriented programming language.
As a further illustration that the presented \ac{oof} is practical to use in real life, the framework is used by TNO's StreetWise program for storing real-world scenarios in a database \autocite{elrofai2018scenario}\footnote{See also \url{https://www.tno.nl/streetwise}.}.

In the example, two different actors are considered: the ego vehicle and the pedestrian. 
These are examples of traffic participants, but an actor is not necessarily a traffic participant. 
For example, road side units that transmit messages in an infrastructure-to-vehicle communication setting can also be actors.
In this case, the transmission of messages can be considered as an activity.
Another example of an actor is the road surface in case it is important for the scenario to model the changing surface temperature.

%% file: figures/example_overview.tikz
% This file was created by matplotlib2tikz v0.6.17.
\begin{tikzpicture}
\selectcolormodel{gray}
\definecolor{color0}{rgb}{0.8,1,0.8}
\definecolor{color1}{rgb}{1,0.9,0.8}
\definecolor{color2}{rgb}{0,0.4375,0.75}
\definecolor{color3}{rgb}{0.9,0.8,0.7}
\definecolor{color4}{rgb}{0.75,0.75,1}
\definecolor{color5}{rgb}{0.5,0.25,0.125}

\begin{axis}[
xmin=-10, xmax=10,
ymin=-6, ymax=6,
width=\figurewidth,
height=0.60\figurewidth,
tick align=outside,
xticklabel style = {align=center,text width=1},
yticklabel style = {align=right,text width=1},
tick pos=left,
x grid style={white!69.01960784313725!black},
y grid style={white!69.01960784313725!black},
axis background/.style={fill=color0},
ticks=none,
scale only axis
]
\path [draw=white!80.0!black, fill=white!80.0!black] (axis cs:3.4,2.8)
--(axis cs:3.4,2.8)
--(axis cs:3.41736481776669,2.80151922469878)
--(axis cs:3.43420201433257,2.80603073792141)
--(axis cs:3.45,2.81339745962156)
--(axis cs:3.46427876096865,2.8233955556881)
--(axis cs:3.4766044443119,2.83572123903135)
--(axis cs:3.48660254037844,2.85)
--(axis cs:3.49396926207859,2.86579798566743)
--(axis cs:3.49848077530122,2.88263518223331)
--(axis cs:3.5,2.9)
--(axis cs:3.5,2.9)
--(axis cs:6.5,2.9)
--(axis cs:6.5,2.9)
--(axis cs:6.50151922469878,2.88263518223331)
--(axis cs:6.50603073792141,2.86579798566743)
--(axis cs:6.51339745962156,2.85)
--(axis cs:6.5233955556881,2.83572123903135)
--(axis cs:6.53572123903135,2.8233955556881)
--(axis cs:6.55,2.81339745962156)
--(axis cs:6.56579798566743,2.80603073792141)
--(axis cs:6.58263518223331,2.80151922469878)
--(axis cs:6.6,2.8)
--(axis cs:6.6,2.8)
--(axis cs:6.6,-2.8)
--(axis cs:6.6,-2.8)
--(axis cs:6.58263518223331,-2.80151922469878)
--(axis cs:6.56579798566743,-2.80603073792141)
--(axis cs:6.55,-2.81339745962156)
--(axis cs:6.53572123903135,-2.8233955556881)
--(axis cs:6.5233955556881,-2.83572123903135)
--(axis cs:6.51339745962156,-2.85)
--(axis cs:6.50603073792141,-2.86579798566743)
--(axis cs:6.50151922469878,-2.88263518223331)
--(axis cs:6.5,-2.9)
--(axis cs:6.5,-2.9)
--(axis cs:3.5,-2.9)
--(axis cs:3.5,-2.9)
--(axis cs:3.49848077530122,-2.88263518223331)
--(axis cs:3.49396926207859,-2.86579798566743)
--(axis cs:3.48660254037844,-2.85)
--(axis cs:3.4766044443119,-2.83572123903135)
--(axis cs:3.46427876096865,-2.8233955556881)
--(axis cs:3.45,-2.81339745962156)
--(axis cs:3.43420201433257,-2.80603073792141)
--(axis cs:3.41736481776669,-2.80151922469878)
--(axis cs:3.4,-2.8)
--(axis cs:3.4,-2.8)
--cycle;

\path [draw=white!80.0!black, fill=white!80.0!black] (axis cs:-10,2.8)
--(axis cs:3.4,2.8)
--(axis cs:3.4,-2.8)
--(axis cs:-10,-2.8)
--cycle;

\path [draw=color1, fill=color1] (axis cs:3.5,-6)
--(axis cs:3.5,-2.9)
--(axis cs:6.5,-2.9)
--(axis cs:6.5,-6)
--cycle;

\path [draw=white!80.0!black, fill=white!80.0!black] (axis cs:6.6,2.8)
--(axis cs:25,2.8)
--(axis cs:25,-2.8)
--(axis cs:6.6,-2.8)
--cycle;

\path [draw=color1, fill=color1] (axis cs:3.5,2.9)
--(axis cs:3.5,6)
--(axis cs:6.5,6)
--(axis cs:6.5,2.9)
--cycle;

\addplot [semithick, black, forget plot]
table {%
3.4 2.8
3.4 2.8
3.41736481776669 2.80151922469878
3.43420201433257 2.80603073792141
3.45 2.81339745962156
3.46427876096865 2.8233955556881
3.4766044443119 2.83572123903135
3.48660254037844 2.85
3.49396926207859 2.86579798566743
3.49848077530122 2.88263518223331
3.5 2.9
3.5 2.9
};
\addplot [semithick, black, forget plot]
table {%
6.5 2.9
6.5 2.9
6.50151922469878 2.88263518223331
6.50603073792141 2.86579798566743
6.51339745962156 2.85
6.5233955556881 2.83572123903135
6.53572123903135 2.8233955556881
6.55 2.81339745962156
6.56579798566743 2.80603073792141
6.58263518223331 2.80151922469878
6.6 2.8
6.6 2.8
};
\addplot [semithick, black, forget plot]
table {%
6.6 -2.8
6.6 -2.8
6.58263518223331 -2.80151922469878
6.56579798566743 -2.80603073792141
6.55 -2.81339745962156
6.53572123903135 -2.8233955556881
6.5233955556881 -2.83572123903135
6.51339745962156 -2.85
6.50603073792141 -2.86579798566743
6.50151922469878 -2.88263518223331
6.5 -2.9
6.5 -2.9
};
\addplot [semithick, black, forget plot]
table {%
3.5 -2.9
3.5 -2.9
3.49848077530122 -2.88263518223331
3.49396926207859 -2.86579798566743
3.48660254037844 -2.85
3.4766044443119 -2.83572123903135
3.46427876096865 -2.8233955556881
3.45 -2.81339745962156
3.43420201433257 -2.80603073792141
3.41736481776669 -2.80151922469878
3.4 -2.8
3.4 -2.8
};
\addplot [semithick, black, forget plot]
table {%
-10 2.8
3.4 2.8
};
\addplot [semithick, black, forget plot]
table {%
3.4 -2.8
-10 -2.8
};
\addplot [semithick, white, forget plot]
table {%
-10 0
-9.44166666666667 0
};
\addplot [semithick, white, forget plot]
table {%
-7.20833333333333 0
-6.09166666666667 0
};
\addplot [semithick, white, forget plot]
table {%
-3.85833333333333 0
-2.74166666666667 0
};
\addplot [semithick, white, forget plot]
table {%
-0.508333333333332 0
0.608333333333335 0
};
\addplot [semithick, white, forget plot]
table {%
2.84166666666667 0
3.4 0
};
\addplot [semithick, color0, forget plot]
table {%
3.5 -6
3.5 -2.9
};
\addplot [semithick, color0, forget plot]
table {%
6.5 -2.9
6.5 -6
};
\addplot [semithick, black, forget plot]
table {%
6.6 2.8
25 2.8
};
\addplot [semithick, black, forget plot]
table {%
25 -2.8
6.6 -2.8
};
\addplot [semithick, white, forget plot]
table {%
6.6 0
7.11111111111111 0
};
\addplot [semithick, white, forget plot]
table {%
9.15555555555556 0
10.1777777777778 0
};
\addplot [semithick, white, forget plot]
table {%
12.2222222222222 0
13.2444444444444 0
};
\addplot [semithick, white, forget plot]
table {%
15.2888888888889 0
16.3111111111111 0
};
\addplot [semithick, white, forget plot]
table {%
18.3555555555556 0
19.3777777777778 0
};
\addplot [semithick, white, forget plot]
table {%
21.4222222222222 0
22.4444444444444 0
};
\addplot [semithick, white, forget plot]
table {%
24.4888888888889 0
25 0
};
\addplot [semithick, color0, forget plot]
table {%
3.5 2.9
3.5 6
};
\addplot [semithick, color0, forget plot]
table {%
6.5 6
6.5 2.9
};
\path [draw=white, fill=white] (axis cs:6.6,2.71875)
--(axis cs:6.6,2.35625)
--(axis cs:3.4,2.35625)
--(axis cs:3.4,2.71875)
--cycle;

\path [draw=white, fill=white] (axis cs:6.6,1.99375)
--(axis cs:6.6,1.63125)
--(axis cs:3.4,1.63125)
--(axis cs:3.4,1.99375)
--cycle;

\path [draw=white, fill=white] (axis cs:6.6,1.26875)
--(axis cs:6.6,0.90625)
--(axis cs:3.4,0.90625)
--(axis cs:3.4,1.26875)
--cycle;

\path [draw=white, fill=white] (axis cs:6.6,0.54375)
--(axis cs:6.6,0.18125)
--(axis cs:3.4,0.18125)
--(axis cs:3.4,0.54375)
--cycle;

\path [draw=white, fill=white] (axis cs:6.6,-0.18125)
--(axis cs:6.6,-0.54375)
--(axis cs:3.4,-0.54375)
--(axis cs:3.4,-0.18125)
--cycle;

\path [draw=white, fill=white] (axis cs:6.6,-0.90625)
--(axis cs:6.6,-1.26875)
--(axis cs:3.4,-1.26875)
--(axis cs:3.4,-0.90625)
--cycle;

\path [draw=white, fill=white] (axis cs:6.6,-1.63125)
--(axis cs:6.6,-1.99375)
--(axis cs:3.4,-1.99375)
--(axis cs:3.4,-1.63125)
--cycle;

\path [draw=white, fill=white] (axis cs:6.6,-2.35625)
--(axis cs:6.6,-2.71875)
--(axis cs:3.4,-2.71875)
--(axis cs:3.4,-2.35625)
--cycle;

\path [draw=black, fill=color2] (axis cs:-7.25,-1.39598214285714)
--(axis cs:-7.24189189189189,-0.873660714285714)
--(axis cs:-7.18513513513513,-0.672767857142857)
--(axis cs:-7.10405405405405,-0.600446428571429)
--(axis cs:-6.97432432432432,-0.552232142857143)
--(axis cs:-6.51216216216216,-0.495982142857143)
--(axis cs:-3.09864864864865,-0.536160714285714)
--(axis cs:-3.00945945945946,-0.592410714285714)
--(axis cs:-2.87162162162162,-0.769196428571428)
--(axis cs:-2.81486486486487,-0.962053571428571)
--(axis cs:-2.75,-1.39598214285714)
--(axis cs:-2.75,-1.40401785714286)
--(axis cs:-2.81486486486487,-1.83794642857143)
--(axis cs:-2.87162162162162,-2.03080357142857)
--(axis cs:-3.00945945945946,-2.20758928571429)
--(axis cs:-3.09864864864865,-2.26383928571429)
--(axis cs:-6.51216216216216,-2.30401785714286)
--(axis cs:-6.97432432432432,-2.24776785714286)
--(axis cs:-7.10405405405405,-2.19955357142857)
--(axis cs:-7.18513513513513,-2.12723214285714)
--(axis cs:-7.24189189189189,-1.92633928571429)
--(axis cs:-7.25,-1.40401785714286)
--cycle;

\path [draw=black, fill=color2] (axis cs:-4.33108108108108,-0.568303571428571)
--(axis cs:-4.33918918918919,-0.367410714285714)
--(axis cs:-4.33108108108108,-0.319196428571428)
--(axis cs:-4.29864864864865,-0.343303571428571)
--(axis cs:-4.25,-0.552232142857143)
--cycle;

\path [draw=black, fill=color2] (axis cs:-4.33108108108108,-2.23169642857143)
--(axis cs:-4.33918918918919,-2.43258928571429)
--(axis cs:-4.33108108108108,-2.48080357142857)
--(axis cs:-4.29864864864865,-2.45669642857143)
--(axis cs:-4.25,-2.24776785714286)
--cycle;

\path [draw=black, fill=white] (axis cs:-6.71486486486486,-1.39598214285714)
--(axis cs:-6.69864864864865,-1.02633928571429)
--(axis cs:-6.65,-0.801339285714286)
--(axis cs:-6.58513513513514,-0.680803571428571)
--(axis cs:-6.52837837837838,-0.672767857142857)
--(axis cs:-6.01756756756757,-0.809375)
--(axis cs:-6.06621621621622,-0.945982142857143)
--(axis cs:-6.07432432432432,-1.15491071428571)
--(axis cs:-6.07432432432432,-1.64508928571429)
--(axis cs:-6.06621621621622,-1.85401785714286)
--(axis cs:-6.01756756756757,-1.990625)
--(axis cs:-6.52837837837838,-2.12723214285714)
--(axis cs:-6.58513513513514,-2.11919642857143)
--(axis cs:-6.65,-1.99866071428571)
--(axis cs:-6.69864864864865,-1.77366071428571)
--(axis cs:-6.71486486486486,-1.40401785714286)
--cycle;

\path [draw=black, fill=white] (axis cs:-3.91756756756757,-1.39598214285714)
--(axis cs:-3.94189189189189,-0.962053571428571)
--(axis cs:-4.03108108108108,-0.688839285714286)
--(axis cs:-4.08783783783784,-0.632589285714286)
--(axis cs:-4.61486486486486,-0.841517857142857)
--(axis cs:-4.56621621621622,-1.034375)
--(axis cs:-4.55,-1.203125)
--(axis cs:-4.55,-1.596875)
--(axis cs:-4.56621621621622,-1.765625)
--(axis cs:-4.61486486486486,-1.95848214285714)
--(axis cs:-4.08783783783784,-2.16741071428571)
--(axis cs:-4.03108108108108,-2.11116071428571)
--(axis cs:-3.94189189189189,-1.83794642857143)
--(axis cs:-3.91756756756757,-1.40401785714286)
--cycle;

\path [draw=black, fill=white] (axis cs:-6.02567567567568,-0.568303571428571)
--(axis cs:-5.81486486486487,-0.568303571428571)
--(axis cs:-5.81486486486487,-0.720982142857143)
--(axis cs:-5.92027027027027,-0.672767857142857)
--cycle;

\path [draw=black, fill=white] (axis cs:-6.02567567567568,-2.23169642857143)
--(axis cs:-5.81486486486487,-2.23169642857143)
--(axis cs:-5.81486486486487,-2.07901785714286)
--(axis cs:-5.92027027027027,-2.12723214285714)
--cycle;

\path [draw=black, fill=white] (axis cs:-5.76621621621622,-0.737053571428571)
--(axis cs:-5.76621621621622,-0.608482142857143)
--(axis cs:-5.73378378378378,-0.576339285714286)
--(axis cs:-5.20675675675676,-0.576339285714286)
--(axis cs:-5.18243243243243,-0.616517857142857)
--(axis cs:-5.23108108108108,-0.761160714285714)
--(axis cs:-5.30405405405405,-0.793303571428571)
--(axis cs:-5.57162162162162,-0.777232142857143)
--cycle;

\path [draw=black, fill=white] (axis cs:-5.76621621621622,-2.06294642857143)
--(axis cs:-5.76621621621622,-2.19151785714286)
--(axis cs:-5.73378378378378,-2.22366071428571)
--(axis cs:-5.20675675675676,-2.22366071428571)
--(axis cs:-5.18243243243243,-2.18348214285714)
--(axis cs:-5.23108108108108,-2.03883928571429)
--(axis cs:-5.30405405405405,-2.00669642857143)
--(axis cs:-5.57162162162162,-2.02276785714286)
--cycle;

\path [draw=black, fill=white] (axis cs:-5.15,-0.817410714285714)
--(axis cs:-5.02027027027027,-0.568303571428571)
--(axis cs:-4.28243243243243,-0.568303571428571)
--(axis cs:-4.29054054054054,-0.616517857142857)
--(axis cs:-4.70405405405405,-0.793303571428571)
--cycle;

\path [draw=black, fill=white] (axis cs:-5.15,-1.98258928571429)
--(axis cs:-5.02027027027027,-2.23169642857143)
--(axis cs:-4.28243243243243,-2.23169642857143)
--(axis cs:-4.29054054054054,-2.18348214285714)
--(axis cs:-4.70405405405405,-2.00669642857143)
--cycle;

\path [draw=black, fill=white] (axis cs:-3.2527027027027,-0.552232142857143)
--(axis cs:-3.09054054054054,-0.560267857142857)
--(axis cs:-2.96891891891892,-0.680803571428571)
--(axis cs:-2.91216216216216,-0.777232142857143)
--(axis cs:-2.89594594594595,-0.873660714285714)
--(axis cs:-2.89594594594595,-1.04241071428571)
--(axis cs:-2.98513513513514,-0.889732142857143)
--cycle;

\path [draw=black, fill=white] (axis cs:-3.2527027027027,-2.24776785714286)
--(axis cs:-3.09054054054054,-2.23973214285714)
--(axis cs:-2.96891891891892,-2.11919642857143)
--(axis cs:-2.91216216216216,-2.02276785714286)
--(axis cs:-2.89594594594595,-1.92633928571429)
--(axis cs:-2.89594594594595,-1.75758928571429)
--(axis cs:-2.98513513513514,-1.91026785714286)
--cycle;

\path [draw=blue, fill=blue] (axis cs:4.8375796178344,-4.96449704142012)
--(axis cs:4.8343949044586,-4.92603550295858)
--(axis cs:4.85668789808917,-4.87869822485207)
--(axis cs:4.84713375796178,-4.8491124260355)
--(axis cs:4.74522292993631,-4.88165680473373)
--(axis cs:4.68152866242038,-4.94378698224852)
--(axis cs:4.65286624203822,-4.93195266272189)
--(axis cs:4.60828025477707,-4.94970414201183)
--(axis cs:4.57643312101911,-4.96745562130178)
--(axis cs:4.50955414012739,-5.03550295857988)
--(axis cs:4.5,-5.12426035502959)
--(axis cs:4.51273885350319,-5.17751479289941)
--(axis cs:4.5796178343949,-5.25443786982249)
--(axis cs:4.73566878980892,-5.35207100591716)
--(axis cs:4.82484076433121,-5.38757396449704)
--(axis cs:5.10509554140127,-5.43786982248521)
--(axis cs:5.1656050955414,-5.43786982248521)
--(axis cs:5.24203821656051,-5.4112426035503)
--(axis cs:5.30891719745223,-5.34615384615385)
--(axis cs:5.37261146496815,-5.31065088757396)
--(axis cs:5.4171974522293,-5.30769230769231)
--(axis cs:5.47770700636943,-5.23668639053254)
--(axis cs:5.4968152866242,-5.16272189349112)
--(axis cs:5.43949044585987,-5.06213017751479)
--(axis cs:5.42993630573248,-4.9792899408284)
--(axis cs:5.39808917197452,-4.90828402366864)
--(axis cs:5.35350318471338,-4.85207100591716)
--(axis cs:5.28662420382166,-4.80473372781065)
--(axis cs:5.19426751592357,-4.78698224852071)
--(axis cs:5.20063694267516,-4.86094674556213)
--(axis cs:5.18152866242038,-4.89644970414201)
--(axis cs:5.1624203821656,-4.90236686390533)
--(axis cs:5.14968152866242,-4.92603550295858)
--(axis cs:5.14649681528662,-4.95266272189349)
--(axis cs:5.15605095541401,-5.02662721893491)
--(axis cs:5.17197452229299,-5.13609467455621)
--(axis cs:5.15605095541401,-5.20414201183432)
--(axis cs:5.11146496815287,-5.24852071005917)
--(axis cs:5.01273885350319,-5.28698224852071)
--(axis cs:4.95222929936306,-5.28698224852071)
--(axis cs:4.85668789808917,-5.25739644970414)
--(axis cs:4.82484076433121,-5.22781065088757)
--(axis cs:4.79936305732484,-5.14792899408284)
--(axis cs:4.80254777070064,-5.05029585798817)
--(axis cs:4.81210191082803,-5.00591715976331)
--cycle;

\path [draw=black, fill=black] (axis cs:4.82484076433121,-5.38757396449704)
--(axis cs:4.9203821656051,-5.47633136094675)
--(axis cs:5.06687898089172,-5.49704142011834)
--(axis cs:5.10191082802548,-5.47633136094675)
--(axis cs:5.10509554140127,-5.43786982248521)
--cycle;

\path [draw=black, fill=black] (axis cs:5.04140127388535,-4.86686390532544)
--(axis cs:5.10191082802548,-4.87278106508876)
--(axis cs:5.14331210191083,-4.88757396449704)
--(axis cs:5.18152866242038,-4.89644970414201)
--(axis cs:5.20063694267516,-4.86094674556213)
--(axis cs:5.19426751592357,-4.78698224852071)
--(axis cs:5.28662420382166,-4.80473372781065)
--(axis cs:5.31528662420382,-4.62130177514793)
--(axis cs:5.30573248407643,-4.60059171597633)
--(axis cs:5.22611464968153,-4.57396449704142)
--(axis cs:5.18789808917197,-4.57988165680473)
--(axis cs:5.1656050955414,-4.70118343195266)
--(axis cs:5.12738853503185,-4.77218934911243)
--cycle;

\path [draw=color3, fill=color3] (axis cs:4.65286624203822,-4.93195266272189)
--(axis cs:4.64012738853503,-4.9112426035503)
--(axis cs:4.60191082802548,-4.88461538461539)
--(axis cs:4.57643312101911,-4.9112426035503)
--(axis cs:4.57643312101911,-4.96745562130178)
--(axis cs:4.60828025477707,-4.94970414201183)
--cycle;

\path [draw=color3, fill=color3] (axis cs:4.87261146496815,-4.94082840236686)
--(axis cs:4.96178343949045,-4.90236686390533)
--(axis cs:5,-4.89940828402367)
--(axis cs:5.05095541401274,-4.91420118343195)
--(axis cs:5.07006369426752,-4.93491124260355)
--(axis cs:5.0828025477707,-4.96449704142012)
--(axis cs:5.09235668789809,-4.97337278106509)
--(axis cs:5.09872611464968,-4.93786982248521)
--(axis cs:5.10509554140127,-4.92603550295858)
--(axis cs:5.07643312101911,-4.89644970414201)
--(axis cs:5.04140127388535,-4.86686390532544)
--(axis cs:4.97770700636943,-4.85798816568047)
--(axis cs:4.92356687898089,-4.8698224852071)
--(axis cs:4.87261146496815,-4.90532544378698)
--cycle;

\path [draw=color4, fill=color4] (axis cs:4.8375796178344,-4.96449704142012)
--(axis cs:4.87261146496815,-4.94082840236686)
--(axis cs:4.87261146496815,-4.90532544378698)
--(axis cs:4.92356687898089,-4.8698224852071)
--(axis cs:4.89171974522293,-4.8698224852071)
--(axis cs:4.85668789808917,-4.87869822485207)
--(axis cs:4.8343949044586,-4.92603550295858)
--cycle;

\path [draw=color4, fill=color4] (axis cs:5.04140127388535,-4.86686390532544)
--(axis cs:5.07643312101911,-4.89644970414201)
--(axis cs:5.10509554140127,-4.92603550295858)
--(axis cs:5.14649681528662,-4.95266272189349)
--(axis cs:5.14968152866242,-4.92603550295858)
--(axis cs:5.1624203821656,-4.90236686390533)
--(axis cs:5.18152866242038,-4.89644970414201)
--(axis cs:5.14331210191083,-4.88757396449704)
--(axis cs:5.10191082802548,-4.87278106508876)
--cycle;

\path [draw=white!10.0!black, fill=white!10.0!black] (axis cs:4.8375796178344,-4.96449704142012)
--(axis cs:4.81210191082803,-5.00591715976331)
--(axis cs:4.80254777070064,-5.05029585798817)
--(axis cs:4.79936305732484,-5.14792899408284)
--(axis cs:4.82484076433121,-5.22781065088757)
--(axis cs:4.85668789808917,-5.25739644970414)
--(axis cs:4.95222929936306,-5.28698224852071)
--(axis cs:5.01273885350319,-5.28698224852071)
--(axis cs:5.11146496815287,-5.24852071005917)
--(axis cs:5.15605095541401,-5.20414201183432)
--(axis cs:5.17197452229299,-5.13609467455621)
--(axis cs:5.15605095541401,-5.02662721893491)
--(axis cs:5.14649681528662,-4.95266272189349)
--(axis cs:5.10509554140127,-4.92603550295858)
--(axis cs:5.09872611464968,-4.93786982248521)
--(axis cs:5.09235668789809,-4.97337278106509)
--(axis cs:5.0828025477707,-4.96449704142012)
--(axis cs:5.07006369426752,-4.93491124260355)
--(axis cs:5.05095541401274,-4.91420118343195)
--(axis cs:5,-4.89940828402367)
--(axis cs:4.96178343949045,-4.90236686390533)
--(axis cs:4.87261146496815,-4.94082840236686)
--cycle;

\path [draw=color5, fill=color5] (axis cs:5.18789808917197,-4.57988165680473)
--(axis cs:5.22611464968153,-4.57396449704142)
--(axis cs:5.30573248407643,-4.60059171597633)
--(axis cs:5.31528662420382,-4.56508875739645)
--(axis cs:5.26114649681529,-4.50295857988166)
--(axis cs:5.22929936305732,-4.49704142011834)
--(axis cs:5.20700636942675,-4.51479289940828)
--cycle;

\addplot [black, forget plot]
table {%
-6.63378378378378 -0.656696428571429
-6.86081081081081 -0.664732142857143
-7.07162162162162 -0.720982142857143
-7.13648648648649 -0.801339285714286
-7.13648648648649 -1.99866071428571
-7.07162162162162 -2.07901785714286
-6.86081081081081 -2.13526785714286
-6.63378378378378 -2.14330357142857
};
\addplot [black, forget plot]
table {%
-3.00945945945946 -0.970089285714286
-2.98513513513514 -0.945982142857143
-2.89594594594595 -1.08258928571429
-2.89594594594595 -1.71741071428571
-2.98513513513514 -1.85401785714286
-3.00945945945946 -1.82991071428571
};
\addplot [black, forget plot]
table {%
-3.26081081081081 -0.664732142857143
-4.0472972972973 -0.632589285714286
-3.00945945945946 -0.970089285714286
-3.00945945945946 -1.82991071428571
-4.0472972972973 -2.16741071428571
-3.26081081081081 -2.13526785714286
};
\addplot [semithick, blue, forget plot]
table {%
-2.75 -1.4
4.5 -1.4
};
\addplot [semithick, blue, forget plot]
table {%
3.75 -0.65
4.5 -1.4
3.75 -2.15
};
\addplot [semithick, blue, forget plot]
table {%
5 -4.85
5 5
};
\addplot [semithick, blue, forget plot]
table {%
4.25 4.25
5 5
5.75 4.25
};
\addplot [semithick, black, forget plot]
table {%
5 0
5 2
};
\addplot [semithick, black, forget plot]
table {%
4.625 1.625
5 2
5.375 1.625
};
\addplot [semithick, black, forget plot]
table {%
5 0
7 0
};
\addplot [semithick, black, forget plot]
table {%
6.625 0.375
7 0
6.625 -0.375
};
\addplot [semithick, black, forget plot]
table {%
5 1
5.1 1
5.1743214109251 0.996926043706003
5.24813513125266 0.98772517306245
5.32093693842672 0.972460239345397
5.39222952228421 0.951235517530571
5.46152588218767 0.924195993989552
5.52835265373337 0.89152637608584
5.59225334231018 0.853449830436276
5.6527914414207 0.810226458456754
5.70955341446317 0.762151519605818
5.76215151960582 0.709553414463167
5.81022645845675 0.652791441420701
5.85344983043628 0.592253342310184
5.89152637608584 0.528352653733366
5.92419599398955 0.461525882187673
5.95123551753057 0.392229522284215
5.9724602393454 0.320936938426719
5.98772517306245 0.248135131252661
5.996926043706 0.174321410925099
6 0.1
6 0
};
\addplot [semithick, black, forget plot]
table {%
6.375 0.375
6 0
5.625 0.375
};
\node at (axis cs:5,2)[
  anchor=south west,
  text=black,
  rotate=0.0
]{ $\north$};
\node at (axis cs:7,0)[
  anchor= west,
  text=black,
  rotate=0.0
]{ $\east$};
\node at (axis cs:5.7,0.7)[
  anchor=south west,
  text=black,
  rotate=0.0
]{ $\head$};
\node at (axis cs:5,0)[
  anchor= east,
  text=black,
  rotate=0.0
]{ $\origin$};
\end{axis}

\end{tikzpicture}

%% file: figures/example_qualitative.tikz
%\newlength\objectwidth\setlength{\objectwidth}{24em}
%\tikzstyle{object}=[draw, text width=\objectwidth-.5em, align=left, line width=1pt, minimum width=\objectwidth, anchor=north west, node distance=3pt]
%\resizebox{\textwidth}{!}{%
\begin{tikzpicture}
\tikzstyle{every node}=[font=\footnotesize]

\node[object, fill=scenariocategory](scenario category){{\bfseries Crossing pedestrian::Scenario category}\\
	description: A pedestrian is crossing the road \\
	\leavevmode\phantom{description: }on a zebra crossing in front of the \\
	\leavevmode\phantom{description: }ego vehicle\\
	physical element: Pedestrian crossing\\
	\leavevmode\phantom{physical element: }qualitative\\
	actors: Ego qualitative, Pedestrian qualitative\\
	activities: Braking, Stationary, Accelerating, \\
	\leavevmode\phantom{activities: }Walking straight\\
	acts: (Ego qualitative, Braking), \\
	\leavevmode\phantom{acts: }(Ego qualitative, Stationary), \\
	\leavevmode\phantom{acts: }(Ego qualitative, Accelerating), \\
	\leavevmode\phantom{acts: }(Pedestrian qualitative, Walking straight)\\
	tags:};

\node[object, fill=category, right=of scenario category.north east, anchor=north west](ego qualitative){{\bfseries Ego qualitative::Actor category}\\
	type: Vehicle\\
	tags: Ego vehicle, Passenger car};

\node[object, fill=category, below=of ego qualitative.south](pedestrian qualitative){{\bfseries Pedestrian qualitative::Actor category}\\
	type: Pedestrian\\
	tags: Pedestrian};

\node[object, fill=category, below=of pedestrian qualitative.south](braking){{\bfseries Braking::Activity category}\\
	model: Sinusoidal\\
	state variable: Speed ($\egospeed$)\\
	tags: Braking};

\node[object, fill=category, below=of braking.south](stationary){{\bfseries Stationary::Activity category}\\
	model: Constant\\
	state variable: Speed ($\egospeed$)\\
	tags: Stationary};

\node[object, fill=category, right=of ego qualitative.north east, anchor=north west](accelerating){{\bfseries Accelerating::Activity category}\\
	model: Linear\\
	state variable: Speed ($\egospeed$)\\
	tags: Accelerating};

%\node[object, fill=category, below=of accelerating.south](straight){{\bfseries Driving straight::Activity category}\\
%	model: Linear\\
%	state variable: Lateral position\\
%	tags: Driving straight};

\node[object, fill=category, below=of accelerating.south](walking){{\bfseries Walking straight::Activity category}\\
	model: Linear\\
	state variable: Position ($\pednorth$)\\
	tags: Walking straight};

\node[object, fill=category, below=of walking.south](static physical){{\bfseries Pedestrian crossing qualitative::Physical element category}\\
	description: Straight road with two lanes and a\\
	\leavevmode\phantom{description: }pedestrian crossing\\
	tags: Non-signalized zebra crossing};

\end{tikzpicture}
%}

%% file: figures/example_quantitative.tikz
%\newlength\objectwidth\setlength{\objectwidth}{24em}
%\tikzstyle{object}=[draw, text width=\objectwidth-.5em, align=left, line width=1pt, minimum width=\objectwidth, anchor=north west, node distance=3pt]
%\resizebox{\textwidth}{!}{%
\begin{tikzpicture}
\tikzstyle{every node}=[font=\footnotesize]

\node[object, fill=scenario](scenario){{\bfseries Ego brakes for pedestrian::Scenario}\\
	physical element: Pedestrian crossing\\
	actors: Ego, Pedestrian\\
	activities: Ego braking, Ego stationary, \\
	\leavevmode\phantom{activities: }Ego accelerating, Pedestrian walking\\
	acts: (Ego, Ego braking), \\
	\leavevmode\phantom{acts: }(Ego, Ego stationary), \\
	\leavevmode\phantom{acts: }(Ego, Ego accelerating), \\
	\leavevmode\phantom{acts: }(Pedestrian, Pedestrian walking)\\
	\attrtstart: Start scenario \\
	\attrtend: End scenario};

\node[object, fill=otherclass, below=of scenario.south](ego){{\bfseries Ego::Actor}\\
	actor category: Ego qualitative\\
	properties: \{width=\SI{1.8}{\meter}, length=\SI{4.5}{\meter}\}\\
	initial state vector: $\egoeast=\SI{-20}{\meter}$, \\
	\leavevmode\phantom{initial state vector: }$\egonorth=\SI{-1.5}{\meter}$, \\
	\leavevmode\phantom{initial state vector: }$\egoheading=\ang{90}$\\
	desired state vector:};

\node[object, fill=otherclass, below=of ego.south](pedestrian){{\bfseries Pedestrian::Actor}\\
	actor category: Pedestrian qualitative\\
	properties: \{width=\SI{0.5}{\meter}, color=blue\}\\
	initial state vector: $\pedeast=\SI{0}{\meter}$, $\pedheading=\ang{0}$\\
	desired state vector:};

\node[object, fill=otherclass, right=of scenario.north east, anchor=north west](start){{\bfseries Start scenario::Event}\\
	time: \SI{0}{\second}};

\node[object, fill=otherclass, below=of start.south](end braking){{\bfseries End braking::Event}\\
	time: \SI{4}{\second}};

\node[object, fill=otherclass, below=of end braking.south](start accelerating){{\bfseries Start accelerating::Event}\\
	time: \SI{7}{\second}};

\node[object, fill=otherclass, below=of start accelerating.south](end scenario){{\bfseries End scenario::Event}\\
	time: \SI{12}{\second}};

\node[object, fill=otherclass, below=of end scenario.south](ego braking){{\bfseries Ego braking::Activity}\\
	activity category: Braking \\
	parameters: $\amplitude=\SI{-8}{\meter\per\second}$, $\duration=\SI{4}{\second}$,\\
	\leavevmode\phantom{parameters: } $\stateinit=\SI{8}{\meter\per\second}$, $\timeinit=\SI{0}{\second}$ \\
	\attrtstart: Start scenario \\
	\attrtend: End braking};

\node[object, fill=otherclass, below=of ego braking.south](ego stationary){{\bfseries Ego stationary::Activity}\\
	activity category: Stationary \\
	parameters: $\stateinit=\SI{0}{\meter\per\second}$ \\
	\attrtstart: End braking \\
	\attrtend: Start accelerating};

\node[object, fill=otherclass, right=of start.north east, anchor=north west](ego accelerating){{\bfseries Ego accelerating::Activity}\\
	activity category: Accelerating \\
	parameters: $\slope=\SI{1.5}{\meter\per\second\squared}$, $\stateinit=\SI{0}{\meter\per\second}$,\\
	\leavevmode\phantom{parameters: } $\timeinit=\SI{7}{\second}$ \\
	\attrtstart: Start accelerating \\
	\attrtend: End scenario};

%\node[object, fill=category, below=of accelerating.south](straight){{\bfseries Driving straight::Activity category}\\
%	model: Linear\\
%	state: Lateral position\\
%	tags: Driving straight};

\node[object, fill=otherclass, below=of ego accelerating.south](pedestrian walking){{\bfseries Pedestrian walking::Activity}\\
	activity category: Walking \\
	parameters: $\slope=\SI{1}{\meter\per\second}$, $\stateinit=\SI{-6}{\meter}$\\
	\leavevmode\phantom{parameters: } $\timeinit=\SI{0}{\second}$ \\
	\attrtstart: Start scenario \\
	\attrtend: End scenario};

\node[object, fill=otherclass, below=of pedestrian walking.south](ped crossing){{\bfseries Pedestrian crossing::Physical element}\\
	physical element category: pedestrian crossing\\
	\leavevmode\phantom{physical element category: }qualitative\\
	properties: \{road: \{lanes: 2, lanewidth: \SI{3}{\meter}, \\
	\leavevmode\phantom{properties: \{road: \{}xy: [(-60, 0), (60, 0)]\}, \\
	\leavevmode\phantom{properties: \{}footway: \{width: \SI{3}{\meter}, \\
	\leavevmode\phantom{properties: \{footway: \{}xy: [(0, 6), (0, -6)]\}\}};

\end{tikzpicture}
%}

%% file: figures/example_test_case.tikz
%\newlength\objectwidth\setlength{\objectwidth}{24em}
%\tikzstyle{object}=[draw, text width=\objectwidth-.5em, align=left, line width=1pt, minimum width=\objectwidth, anchor=north west, node distance=3pt]
%\resizebox{\textwidth}{!}{%
\begin{tikzpicture}
\tikzstyle{every node}=[font=\footnotesize]

\node[object, fill=scenariocategory](scenario category){{\bfseries Crossing pedestrian::Scenario category}\\
	description: A pedestrian is crossing the road \\
	\leavevmode\phantom{description: }on a zebra crossing in front of the \\
	\leavevmode\phantom{description: }ego vehicle\\
	physical element: Pedestrian crossing \\
	\leavevmode\phantom{physical element: }qualitative\\
	actors: Ego qualitative, Pedestrian qualitative\\
	activities: Walking straight\\
	acts: (Pedestrian qualitative, Walking straight)};

\node[object, fill=scenario, below=of scenario category.south](scenario){{\bfseries Ego must brake for pedestrian::Scenario}\\
	physical element: Pedestrian crossing\\
	actors: Ego, Pedestrian\\
	activities: Pedestrian walking\\
	acts: (Pedestrian, Pedestrian walking)\\
	events: Start walking, Stop walking\\
	\attrtstart: Start scenario \\
	\attrtend: End scenario};

\node[object, fill=otherclass, right=of scenario category.north east, anchor=north west](ego){{\bfseries Ego::Actor}\\
	actor category: Ego qualitative\\
	properties: \{width=\SI{1.8}{\meter}, length=\SI{4.5}{\meter}\}\\
	initial state vector: $\egoeast=\SI{-60}{\meter}$, \\
	\leavevmode\phantom{initial state vector: }$\egonorth=\SI{-1.5}{\meter}$, \\
	\leavevmode\phantom{initial state vector: }$\egoheading=\ang{90}$, \\
	\leavevmode\phantom{initial state vector: }$\egospeed=\SI{8}{\meter\per\second}$\\
	desired state vector: $\egoeast=\SI{20}{\meter}$, \\ 
	\leavevmode\phantom{desired state vector: }$\egonorth=\SI{-1.5}{\meter}$, \\ 
	\leavevmode\phantom{desired state vector: }$\egoheading=\ang{90}$, \\ 
	\leavevmode\phantom{desired state vector: }$\egospeed=\SI{8}{\meter\per\second}$};

\node[object, fill=otherclass, right=of ego.north east, anchor=north west](start walking){{\bfseries Start walking::Event}\\
	condition: $|\egoeast/\egospeed| \leq \SI{2.5}{\second}$};

\node[object, fill=otherclass, below=of start walking.south](end walking){{\bfseries End walking::Event}\\
	contition: $\pednorth = \SI{6}{\meter}$};

\node[object, fill=otherclass, below=of end walking.south](walking){{\bfseries Pedestrian walking::Activity}\\
	activity category: Walking\\
	parameters: $\slope=\SI{1}{\meter\per\second}$, $\stateinit=\SI{-6}{\meter}$,\\
	\leavevmode\phantom{parameters: }$\timeinit=\text{At Start walking}$\\
	\attrtstart: Start walking \\
	\attrtend: End walking};

\node[object, fill=otherclass, below=of walking.south](end){{\bfseries End scenario::Event}\\
	condition: $\egoeast \geq \SI{20}{\meter}$ OR \textit{collision} OR\\
	\leavevmode\phantom{condition: }$\egonorth \leq \SI{-2}{\meter}$ OR $\egonorth \geq \SI{-1}{\meter}$ \\
	\leavevmode\phantom{condition: }OR $\time > \SI{100}{\second}$.};

\end{tikzpicture}
%}

%% file: sections/conclusion.tex
\section{Conclusions}
\label{sec:conclusion}

The performance assessment of \acp{av} is essential for the legal and public acceptance of \acp{av} as well as for the technology development of \acp{av}. 
Because scenarios are crucial for the assessment, a clear definition of a scenario is required.
In this work, we have proposed a new definition of the concept scenario in the context of the performance assessment \acp{av}. 
 
While our definition is consistent with other definitions from the literature, it is more concrete such that it can directly be implemented using code.
We have further defined the notions of event, activity, and scenario category. 
To formalize the concepts of scenario, event, activity, and scenario category, an \ac{oof} has been proposed. 
Using the proposed framework, it is possible to describe a scenario in both a qualitative and quantitative manner. 
The framework, represented using class diagrams, can be directly translated into a class structure for an object-oriented software implementation. 
This allows us to translate scenarios into code, such that both domain experts and software programs, such as simulation tools, are able to understand the content of the scenarios. 
To demonstrate this, we have made our implementation in the coding language Python publicly available.

The \ac{oof} has been illustrated with an example of an urban scenario with a pedestrian crossing. 
We have also demonstrated how this particular scenario can be used to define a test scenario using the proposed framework.
In the publicly available\footnote{\url{https://github.com/ErwindeGelder/ScenarioDomainModel}} coding implementation of the presented \ac{oof}, we have shown how to use the proposed \ac{oof} from a real application's perspective.

The presented framework is applicable for scenario mining \autocite{paardekooper2019dataset6000km, degelder2020scenariomining} and scenario-based assessment \autocite{elrofai2018scenario, putz2017pegasus} and, therefore, this framework provides a step towards scenario-based performance assessment of \acp{av}. 
The next step is to define scenarios and scenario categories\footnote{As a starting point, the 67 scenario categories in \autocite{degelder2019scenariocategories} can be used.} that are relevant for \iac{av} in a specific deployment area. 
Future work also includes creating an ontology for scenarios for the assessment of \acp{av}. The presented \ac{oof} could be a good starting point for this \autocite{siricharoen2009ontology}. 
An ontology allows, among others, to add properties to relationships that enable automated reasoning. 
In this way, an ontology enables automated classification of scenarios, thereby helping to overcome problems of data ambiguity \autocite{OpenSCENARIO2}.

%% file: sections/nomenclature.tex
\section{Nomenclature}
\label{sec:nomenclature}

For the definition of \emph{scenario}, several notions are adopted from the literature. 
In this section, the concepts of \emph{ego vehicle}, \emph{physical element}, \emph{actor}, \emph{static environment}, \emph{dynamic environment}, \emph{act}, \emph{state variable}, \emph{state vector}, \emph{model}, and \emph{mode}, which are adopted from literature, are detailed.

\subsection{Ego vehicle}
\label{sec:ego vehicle}

The ego vehicle is the main subject of a scenario. 
In particular, the ego vehicle refers to the vehicle that is perceiving the world through its sensors (see, e.g., \autocite{Bonnin2014}). 
When performing tests, the ego vehicle also refers to the vehicle that must perform a specific task (see, e.g., \autocite{althoff2017CommonRoad, catapult2018musicc}). 
In this case, the ego vehicle is often referred to as the system under test \autocite{stellet2015taxonomy}, the vehicle under test \autocite{gietelink2006development}, or the host vehicle \autocite{gietelink2006development}.

\subsection{Physical element}
\label{sec:physical element}

A physical element refers to an object that exists in the three-dimensional space.

\subsection{Actor}
\label{sec:actor}

According to \textcite{catapult2018musicc}, ``actors are all dynamic components of a scenario, excluding the ego vehicle itself.'' 
Note that, in contrast to \autocite{catapult2018musicc}, in the current paper, the ego vehicle's driver, and/or automation system are considered as actors, similar to \autocite{geyer2014},  because they have the same properties as another driver or automation system.
While the aforementioned definition of \textcite{catapult2018musicc} provides a good idea of what an actor could be, we use another definition in order to avoid a circular definition: an actor is a dynamic physical element, i.e., a physical element that experiences change.

\begin{remark}
	An actor is also a physical element whereas a physical element is not necessarily an actor.
	For example, a static road sign is considered a physical element, but because it does not change during the course of a scenario, it is not an actor.
\end{remark}

\subsection{Static environment}
\label{sec:static environment}

The static environment refers to the part of the environment that does not change during a scenario. This includes geo-spatially stationary elements \autocite{ulbrich2015},  such as the road network.

\subsection{Dynamic environment}
\label{sec:dynamic environment}

As opposed to the static environment, the dynamic environment refers to the part of the environment that changes during the time frame of a scenario. 
In practice, the dynamic environment mainly consists of the moving actors (other than the ego vehicle) that are relevant to the ego vehicle.
For example, the primary use case of OpenSCENARIO \autocite{openscenario}, a file format for the description of the dynamic content of driving simulations, is to describe ``complex, synchronized maneuvers that involve multiple entities like vehicles, pedestrians, and other traffic participants'' \autocite{openscenario}; so for OpenSCENARIO, these maneuvers represent the dynamic environment.
Roadside units that communicate with vehicles within the communication range \autocite{alsultan2014comprehensive} are also part of the dynamic environment. Furthermore, changing (weather) conditions are part of the dynamic environment.

\begin{remark}
	Note that it might not always be obvious whether an element of the environment belongs to the static or dynamic environment. 
	Most important, however, is that all parts of the environment that are relevant to the assessment of an AV are described in either the static or the dynamic environment.
\end{remark}

\subsection{Act}
\label{sec:act}

We define an act as a combination of an actor and the activity that is performed by the actor or the combination of actors and the activities they are subjected to.
This is in accordance with the use of the term \emph{act} in \autocite{openscenario}.

\subsection{State variable} 
\label{sec:state variable}
\textcite[p.~163]{dorf2011modern} write that ``the state variables describe the present configuration of a system and can be used to determine the future response, given the excitation inputs and the equations describing the dynamics.'' 
In our case, ``the system'' could refer to an actor, a component, or a simulation.
For example, a state variable could be the acceleration of an actor.

\subsection{State vector}
\label{sec:state vector}
A state vector refers to ``the vector containing all $n$ state variables'' \autocite[p.~233]{dorf2011modern}.

\subsection{Model}
\label{sec:model}

A dynamical system is often modeled using a differential equation of the form $\statedot(\time)=\function_{\parameter}(\state(\time), \inputsystem(\time), \time)$ \autocite{norman2011control}, where $\state(\time)$ represents the state vector at time $\time$, $\inputsystem(\time)$ represents an external input vector, and the function $\function_{\parameter}(\cdot)$ is parameterized by $\parameter$.  Note that, technically speaking, $\state(\cdot)$, $\inputsystem(\cdot)$, $\time$, and $\parameter$ are inputs of the function $\function$, but $\parameter$ is assumed to be constant for a certain time interval. For example, the following first-order model is parameterized by $\parameter=(\parametera,\parameterb)$:
\begin{equation}
	\statedot(\time) = \parametera \state(\time) + \parameterb \inputsystem(\time).
\end{equation}

\subsection{Mode}
\label{sec:mode}

In some systems, the behavior of the system may suddenly change abruptly, e.g., due to a sudden change in an input, a model parameter, or the model.
Such a transition is called a mode switch.
In each mode, the behavior of the system is described by a model with a fixed function $f_{\theta}$ and smooth input $u(\cdot)$ \autocite{deschutter2000optimal}.

%% file: sections/acknowledgement.tex
\section*{Acknowledgment}

The research leading to this paper has been partially realized with the Centre of Excellence for Testing and Research of Autonomous Vehicles at NTU (CETRAN), Singapore. 
% Responsibility for the information and view set out in this publication lies entirely with the authors. 

We would like to thank Mark van den Brand and Ludwig Friedmann for providing helpful feedback on earlier versions of this article.